\documentclass[conference]{IEEEtran}
\usepackage{times}

\usepackage[numbers]{natbib}
\usepackage{multicol}

\usepackage{microtype}
\usepackage{graphicx}
\usepackage{subcaption}
\usepackage{booktabs} %
\usepackage{easy-todo}
\usepackage{makecell}
\usepackage[bookmarks=true]{hyperref}
\hypersetup{
  pdftitle={RoboLab: A High-Fidelity Simulation Benchmark for Analysis of Task Generalist Policies},
  pdfauthor={Jenai Xuning Yang, Rishit Dagli, Alex Zook, Hugo Hadfield, Ankit Goyal, Stan Birchfield, Fabio Ramos, Jonathan Tremblay},
  pdfsubject={Robotics; Robot Learning; Simulation Benchmarks},
  pdfkeywords={robot learning; simulation benchmark; generalist policies; vision-language-action models}
}
\usepackage{algorithm}
\usepackage{algorithmic}
\usepackage{amsmath}
\usepackage{amssymb}
\usepackage{mathtools}
\usepackage{amsthm}
\usepackage{xspace}
\usepackage[capitalize,noabbrev]{cleveref}
\usepackage{graphicx}
\usepackage{listings}
\usepackage[dvipsnames, table]{xcolor}
\usepackage{tcolorbox}
\usepackage{comment}
\definecolor{nvidiagreen}{HTML}{76B900}
\usepackage{booktabs}
\usepackage{multirow}
\usepackage{cuted} %
\usepackage{caption} %

\let\labelindent\relax
\usepackage{enumitem} %
\newcommand{\framework}{RoboLab\xspace}
\newcommand{\benchname}{RoboLab-120\xspace}
\newcommand{\numtasks}{120\xspace}
\newcommand{\piZero}{\ensuremath{\pi_0}\xspace}
\newcommand{\piZeroFast}{\ensuremath{\pi_0\text{-FAST}}\xspace}
\newcommand{\piZeroFive}{\ensuremath{\pi_{0.5}}\xspace}
\newcommand{\paliGemma}{PaliGemma\xspace}
\newcommand{\groot}{GR00T N1.6\xspace}

\makeatletter
\renewcommand\paragraph[1]{%
  \vspace{0.5\baselineskip}%
  \noindent\textbf{#1.}~%
}
\makeatother

\begin{document}

\title{\framework: A High-Fidelity Simulation Benchmark for Analysis of Task Generalist Policies}

\author{
  \authorblockN{
    Jenai Xuning Yang\textsuperscript{1},
    Rishit Dagli\textsuperscript{2,4},
    Alex Zook\textsuperscript{1},
    Hugo Hadfield\textsuperscript{1},\\
    Ankit Goyal\textsuperscript{1},
    Stan Birchfield\textsuperscript{1},
    Fabio Ramos\textsuperscript{1,3}, and
    Jonathan Tremblay\textsuperscript{1}
  }
  \authorblockA{\small \textsuperscript{1}NVIDIA, \textsuperscript{2}University of Toronto, \textsuperscript{3}The University of Sydney,
  \textsuperscript{4}Work done during internship at NVIDIA
  }
}

\maketitle

\begin{strip}
    \centering
    \vspace*{-3.5em}
    \includegraphics[width=1\linewidth]{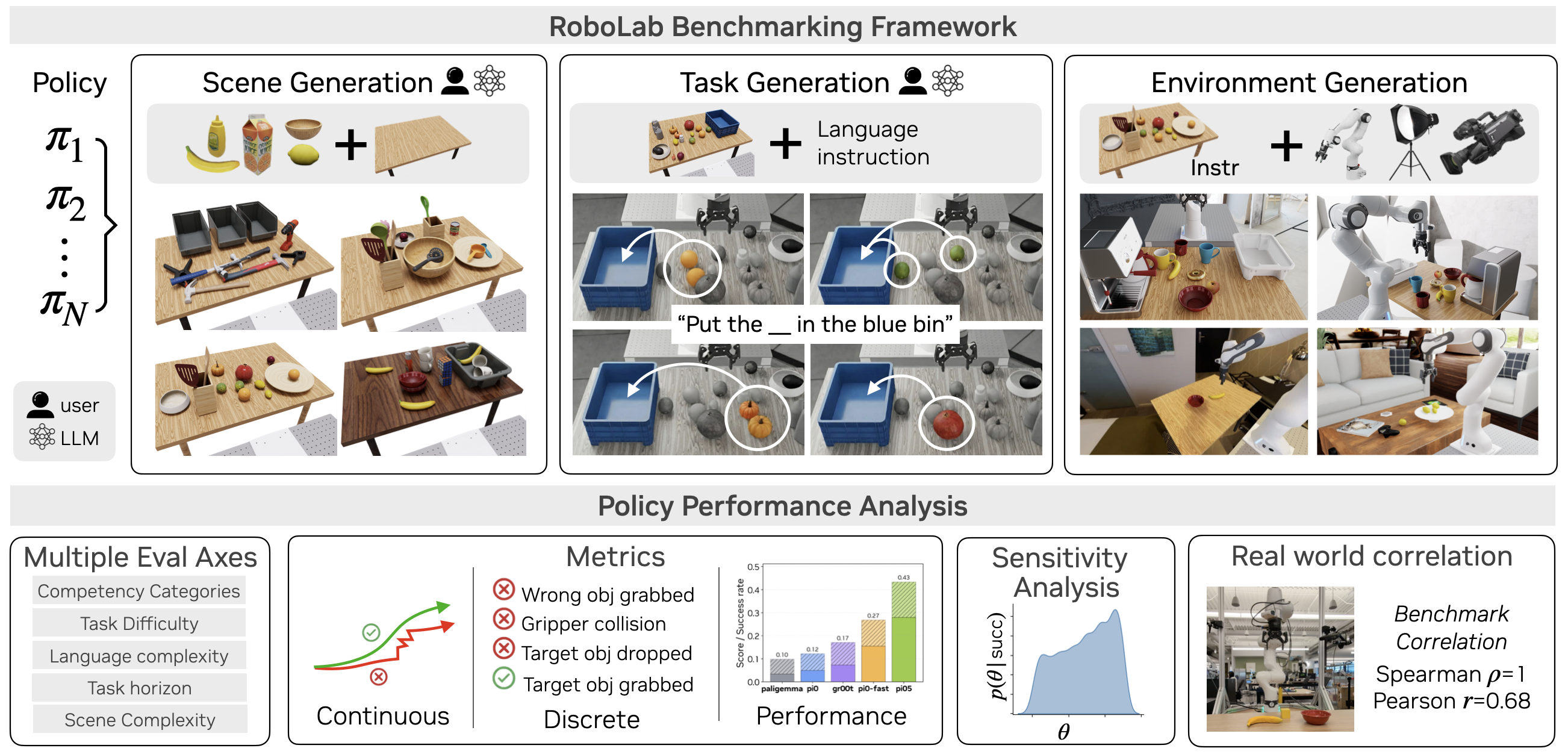}
    \captionof{figure}{\small \textbf{Overview of RoboLab.}
      RoboLab addresses large-scale evaluation of real-world policies via a novel evaluation framework.
      By introducing a streamlined pipeline for generating new scenes and tasks (top row), RoboLab enables rapid extensibility for testing generalization capabilities. The accompanying benchmark, {\emph \benchname}, introduces various evaluation axes and a suite of metrics and analysis tools, and demonstrate benchmark-level correlation with real-world benchmarks (bottom row).
    }
    \label{fig:overview}
\end{strip}

\begin{abstract}
The pursuit of general-purpose robotics has yielded impressive foundation models, yet simulation-based benchmarking remains a bottleneck due to rapid performance saturation and a lack of true generalization testing.
Existing benchmarks often exhibit significant domain overlap between training and evaluation, trivializing success rates and obscuring insights into robustness.
We introduce \framework, a simulation benchmarking framework designed to address these challenges.
Concretely, our framework is designed to answer two questions: (1) to what extent can we understand the performance of a real-world policy by analyzing its behavior in simulation, and (2) which factor most strongly affect policy behavior.
First, \framework enables human-authored and LLM-enabled generation of scenes and tasks in a robot- and policy-agnostic manner within a high-fidelity simulation environment.
We introduce an accompanying \benchname benchmark, consisting of \numtasks tasks categorized into three competency axes: visual, procedural, relational, across three difficulty levels.
Second, we introduce a systematic analysis of real-world policies that quantify both their performance and the sensitivity of their behavior to controlled perturbations, exposing significant performance gap in current state-of-the-art models.
By providing granular metrics and a scalable toolset, \framework offers a scalable framework for evaluating the true generalization capabilities of task-generalist robotic policies. Project website: \url{https://research.nvidia.com/labs/srl/projects/robolab/}.

\end{abstract}
\IEEEpeerreviewmaketitle

\section{Introduction}

\begin{figure*}
    \centering
    \includegraphics[width=\linewidth]{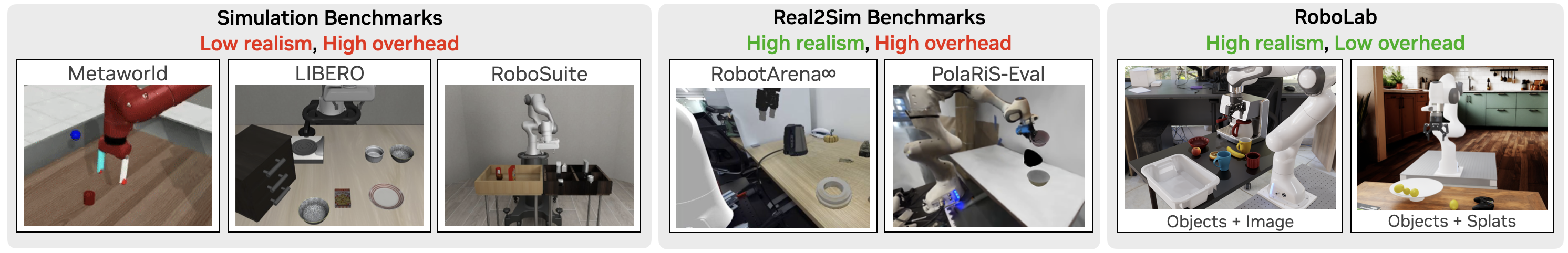}
    \caption{\small Three approaches for robotic benchmarks. {\sc Left:} To date, pure simulation based benchmarks have exhibited low visual quality, creating a large sim2real transfer gap. {\sc Middle:} Real2sim benchmarks address this issue by using techniques to bring real-world visual texture into simulation. However, these environments are extremely costly with reported per-scene generation time of $\sim$1hr~\cite{polaris}.  {\sc Right:} Our approach achieves a high degree of realism with low overhead.}
    \label{fig:benchmark_comparison}
\end{figure*}

The pursuit of generality has been a longstanding challenge in modern robotics.
Recent advances have produced impressive generalist robot policies that demonstrate success in challenging and novel tasks in the real-world.
Despite this progress, benchmarks for evaluating whether these policies are truly task-general has been slow.
Evaluating models in the real world remains prohibitively expensive and logistically expensive, motivating the rise of simulation-based benchmarks as an appealing alternative.

Current robotics benchmarks~\cite{libero,liberopro,rlbench,simpler} face several critical limitations:
(1) a lack of high-fidelity simulation aimed at evaluation of real-world policies;
(2) rapid performance saturation on static task sets; and
(3) a lack of granular analysis regarding policy failure modes.

For instance, popular benchmarks like LIBERO~\cite{libero} often utilize nearly identical environments for both training and evaluation.
When policies are fine-tuned on these simulation-specific demonstrations, the lack of a meaningful task domain gap trivializes the evaluation process and obscures the model's true generalization capabilities.
Many existing platforms have limited realism or are difficult to extend due to using rigid architectures that make it cumbersome to introduce new objects, tasks, or robots (Fig.~\ref{fig:benchmark_comparison}).

To address these limitations, we present \framework (Fig.~\ref{fig:overview}), a benchmarking framework designed for rigorous evaluation of generalist policies trained on real-world data.
Unlike prior benchmarks that rely on PDDL or rigid scene-graph definitions~\cite{libero}, RoboLab introduces an easy-to-use framework that enables human-authored and AI-enabled scene and task generation.
This enables fast creation of hundreds of new tasks and scenes, providing a scalable framework that mitigates benchmark saturation and ensures long-term value.

We introduce the \benchname benchmark, comprising of \numtasks hand-curated diverse pick-and-place tasks, aimed at evaluating the generalization capabilities.
These tasks span varying difficulties (65 simple, 38 moderate, 18 complex) and multiple competency axes (44 relational, 91 visual, and 36 procedural). This benchmark reflects tasks encountered in ``in the wild'' household scenarios. To prevent evaluation of policies overfitted to a particular simulation domain, we evaluate policies trained exclusively on the real-world DROID~\cite{droid} dataset.
Our analysis shows that the results on \benchname achieve benchmark-level correlation with real-world benchmarks \cite{roboarena}, indicating that \framework is a meaningful proxy for real-world evaluation for task-generalism.

Lastly, we introduce a suite of analysis tools that gives insight into the model performance beyond binary success rates and broader understanding of policy performance; including subtask scoring, event tracking, Bayesian-based sensivity analysis of performance given scene variations via Neural Posterior Estimation.

In summary, our contributions are:
\begin{enumerate}
    \item \textbf{RoboLab}: A novel benchmarking platform designed for evaluating real-world robotics policies with a scalable, AI-based workflow capable of procedurally generating over hundreds of unique robot- and policy- agnostic scenes and tasks, built on IsaacLab\cite{mittal2025isaaclab}.

\item \textbf{\benchname Benchmark}: \numtasks tasks diversely represented across three distinct competency axes (visual, procedural, relational), across 3 levels of difficulty, and supported by robustness metrics.

\item \textbf{Policy Analysis}: We introduce a suite of analysis tools that gives insight into the model performance beyond binary success rates and broader understanding of policy performance.
\end{enumerate}

\begin{figure*}
    \centering
    \includegraphics[width=1\linewidth]{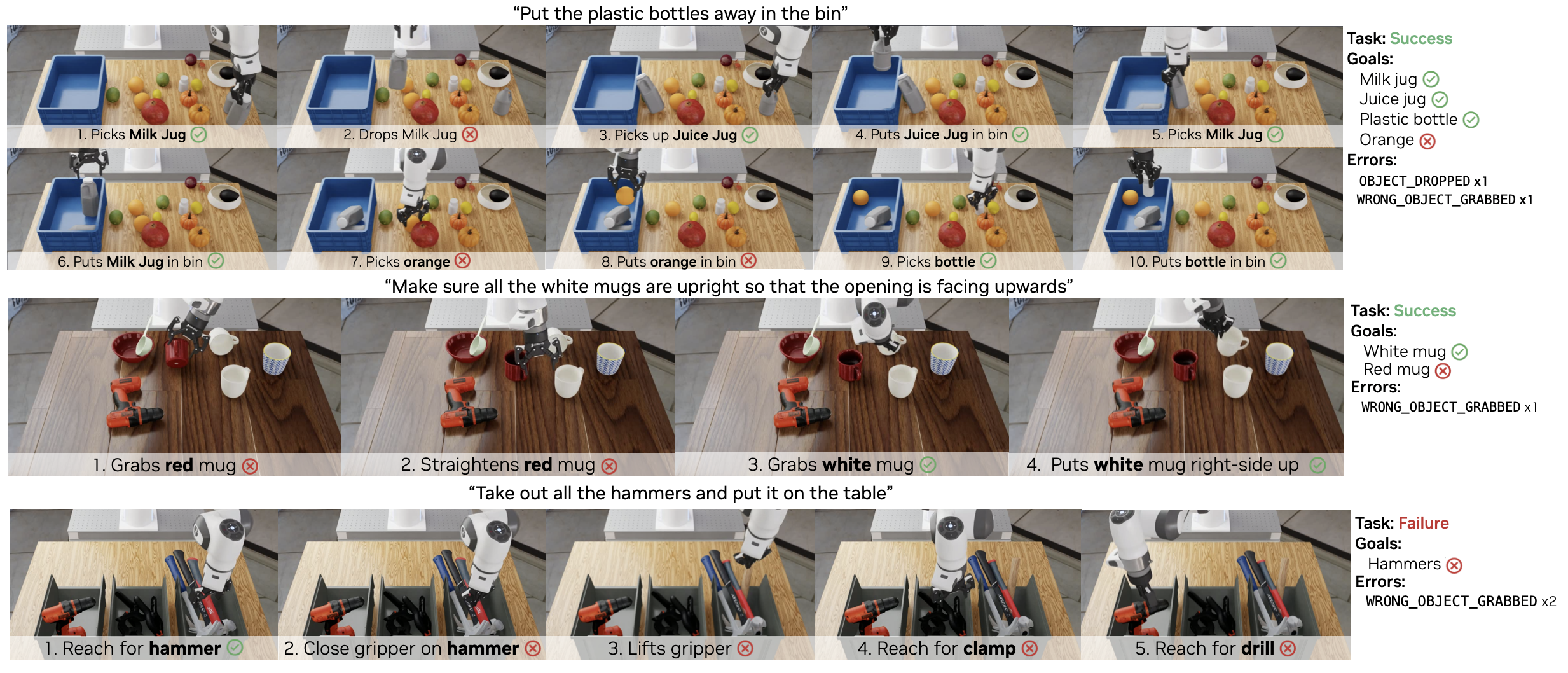}
    \caption{\small \textbf{Task progression of a few tasks, illustrating errors encountered during policy rollout.} Top row: Although the task is successfully completed, errors were encountered during execution: 1) The robot drops the milk jug too early, missing the bin. 2) the robot grasps an orange (wrong object) and puts it in the bin. Mid row: An extraneous object was reoriented before the actual intended object. Final row: Intended objects were attempted unsuccessfully, and the policy tended to two wrong other objects.  }
    \label{fig:task-progression}
\end{figure*}

\section{Related Work}

\paragraph{Simulation-Based Benchmarks}
Simulation provides a scalable and reproducible environment for evaluating robot manipulation policies.
Widely used benchmarks such as RLBench~\cite{rlbench}, MetaWorld~\cite{metaworld}, and robosuite~\cite{robosuite}, ManiSkill2~\cite{gu2023maniskill2}, CALVIN~\cite{mees2022calvin}, LIBERO~\cite{libero},  and BEHAVIOR-1K~\cite{li2024behavior1k}, offer standardized task suites for learning and evaluation in simulation across pre-defined task families and object configurations.
However, in these settings, policies are typically trained and evaluated in the same simulated environments, which encourages overfitting to simulator-specific quirks, leads to rapid benchmark saturation, and makes real-world generalization hard to assess. \cite{liberopro}.
In our setting, policies are instead trained on large-scale real-world data (e.g., DROID~\cite{droid}), while high-fidelity simulation is used only as a controlled evaluation environment. Training and evaluation domains are decoupled and measured performance more closely reflects robustness in the real world.
Some benchmarks also solely focus on perturbations and variations to probe policy robustness (LIBERO~\cite{libero}, REALM~\cite{sedlacek2025realm}); but on a limited task set. RoboLab's task generation and diagnostic analysis allows large-scale evaluation and performance analysis, complementary to existing benchmarks.

\paragraph{Real-to-sim Evaluation}
Recent work have focused on leveraging 3D reconstruction to build photorealistic simulation scenes from real-world videos in order to achieve closer visual alignment between simulation and real world photorealism \cite{simpler,robotarena,polaris,zook2025grs}. These works typically use Gaussian splatting, 3D segmentation, and multi-view inpainting, often operated at a per-scene level, which entails costly optimization and makes it slow to scale beyond a small number of environments  \cite{polaris, real2sim-eval, splatsim} (Appendix~\ref{sec:benchmark-comparison}).
In contrast, our framework produces large-scale, photorealistic scenes and tasks within minutes rather than hours, while preserving sufficient geometric and visual fidelity for policy evaluation, thereby making real-to-sim benchmarking practical at the scale needed for modern generalist robot policies.

\section{\framework}

\begin{figure}[b!]
    \centering
    \includegraphics[width=1\linewidth]{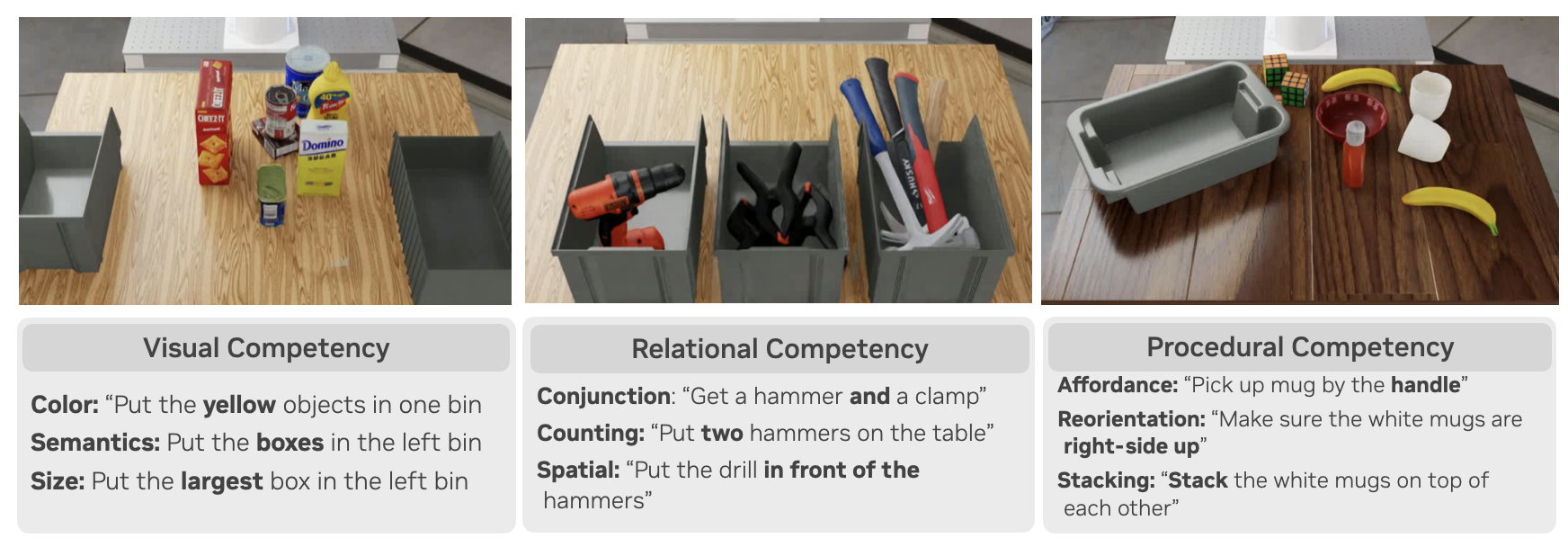}
    \caption{\small Example of language instructions in \benchname.
    }
    \label{fig:rqas}
\end{figure}

\begin{figure*}
    \centering
    \includegraphics[width=1\linewidth]{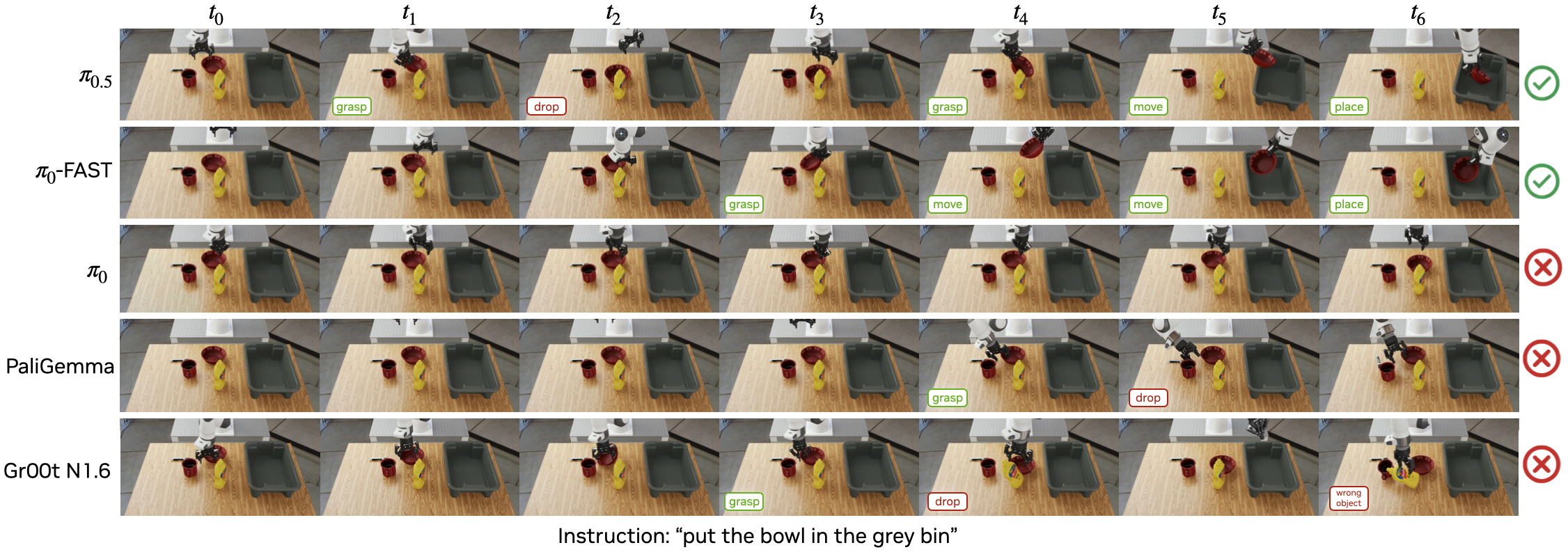}
    \caption{\small 
Comparison of policy performance for bowl-in-bin manipulation. Rows represent distinct policies shown in chronological order (left to right). Successful execution involves grasping the central red bowl and depositing it into the gray bin on the right. Unsuccessful attempts are characterized by aimless arm trajectories and a lack of object interaction. 
    }
    \label{fig:grid}
\end{figure*}

Evaluating real-world, generalist robotics policies in simulation remains a significant challenge. \framework aims to address this with a few design principles: 1) Enable easy and fast scene/task generation; 2) Tasks in \framework are policy- and robot- agnostic, enabling comparison of multiple policy and robot choices when it comes to solving real-world tasks; and 3) generate tasks that are diverse, representative of real-world tasks, and enable a multifaceted analysis of models to understand their generalization capabilities.

\subsection{RoboLab Scene and Task Generation}

\framework introduces a  a user-friendly workflow that mirrors the process of preparing a real-world robot evaluation (Fig.~\ref{fig:overview}):
1) First, create a \textbf{scene} by positioning and orienting objects in a workspace;
2) Then, define a \textbf{task} as language instructions for a goal state in the scene;
3) Lastly, instantiate an \textbf{environment} by selecting a robot, policy, and variations of scene features including camera, lighting, and backgrounds for a task.
\framework enables evaluation of the same tasks across different robot embodiments.
By deferring robot- and experiment-specific binding to runtime, we facilitate systematic evaluation of tasks across robot configurations and policy variants, without having embodiment-specific training.

Formally, define a \emph{scene} $S = \{(b_i, \mathbf{p}_i, \mathbf{q}_i)\}_{i=1}^N$, where $b_i$ represents an object instance selected from the available catalog of objects $\mathcal{B}$ and $\mathbf{p}_i \in \mathbb{R}^3, \mathbf{q}_i \in SO(3)$ denote its position and orientation.
Define a \emph{task} $\mathcal{T}=\{S, l\}$, where $l$ is the \emph{language instruction} to complete in the scene.
Define a policy $\pi\!:\!\mathcal{O}\!\rightarrow\!\mathcal{A}$ where the action space $\mathcal{A} \in \{ \mathcal{A}^{\text{joint}}, \mathcal{A}^{\text{EE}}, \dots \}$ and observation space $\mathcal{O} = (\mathcal{O}^{\text{proprio}}, \mathcal{O}^{\text{rgb}}, \mathcal{O}^{\text{depth}} \cdots)$ is policy dependent.
An environment $E = (\mathcal{T}, \mathcal{R}, \mathcal{O}, \mathcal{A}, \xi)$ consists of a task, robot embodiment $\mathcal{R}$, policy parameters ($\mathcal{A}$, $\mathcal{O}$), and scene variations \(\xi = (\xi^\text{camera}, \xi^\text{light}, \xi^\text{background}, \xi^\text{pose})\).
More details on the specific objects, scenes and tasks in \framework can be found in Appendix~\ref{app:benchmark}. To scale scene and task generation, we introduce an automated workflow, described below:

\subsubsection{Scaling Scene Generation}
\label{sec:scenegen}

We enable scaling scene generation through an automated pipeline that:
1) prompts an LLM to generate a structured scene plan for asset placement;
2) uses a geometric solver and physics simulation to check asset placement validity; and
3) refines the scene if it is not valid.
First, the LLM is prompted with a theme (e.g., ``messy counter'') to generate a structured scene plan consisting of a subset of objects $B \subset \mathcal{B}$ and spatial predicates $\mathcal{P}$ governing the layout.
The LLM is provided with the full catalog of objects $\mathcal{B}$ containing names and bounding box dimensions $\mathbf{d}_i \in \mathbb{R}^3$.
Second, a spatial solver converts the relational predicates $\mathcal{P}$ into valid pose configurations $(\mathbf{p}, \mathbf{q})$).
Objects are processed in dependency order, with support surfaces placed before objects on those surfaces (\cref{alg:spatial}).
To check physical stability, the scene is then forward simulated in Isaac Sim~\cite{NVIDIA_Isaac_Sim} for $300$ steps under gravity.
An object $b_i$ is flagged as \textit{unstable} if it's maximum Euclidean displacement is larger than a threshold (typically $0.02$m).
Third, If any object is unstable, we generate a text error describing the failure (e.g., ``Object `apple' fell off `plate' with displacement 0.15m'').
This feedback is provided to the LLM to refine the scene plan and repeat the process.
Further details, including on the spatial and physical solvers, are provided in Appendix~\ref{sec:scaling-scene-generation}.

\begin{figure*}[t]
    \centering
    \includegraphics[width=1\linewidth]{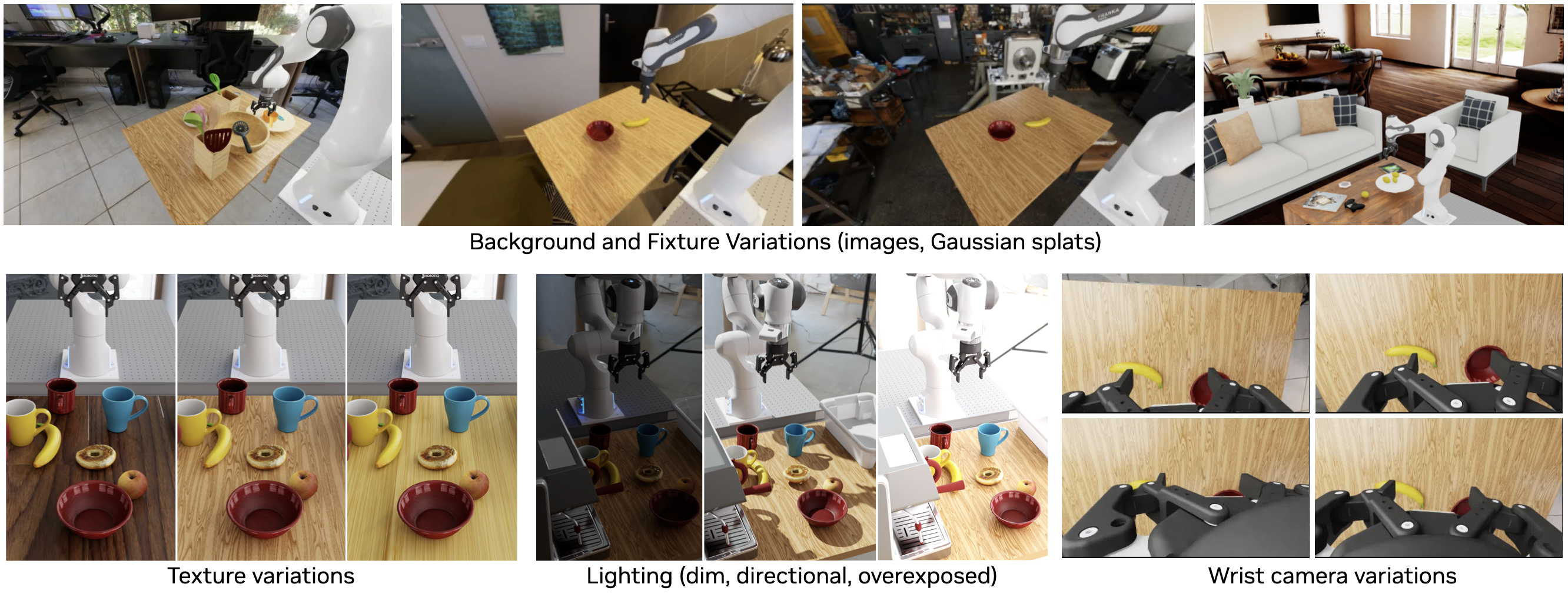}
    \caption{Example scene variations, lighting variations, and camera pose variations in RoboLab. }
    \label{fig:variations}
\end{figure*}

\subsubsection{Scaling Task Generation}
\label{sec:taskgen}

We enable scaling task generation through an automated pipeline that:
1) generates task code from information including the scene and competency axes;
2) validates code syntax;
3) validates asset selections in the scene; and
4) refines the task if it is not valid.
First, we prompt an LLM with detailed task information:
1) the scene object catalog $B_S$ and metadata with dimensions;
2) task examples demonstrating the task structure;
3) the complete predicate library defining sub-task success and termination;
4) Competency-axes language templates with placeholders for objects, spatial verbs, and attributes; and
5) constraints including difficulty levels and physical feasibility requirements (e.g., containment size constraints, stacking stability).
Second, tasks are check for syntax validity as code.
Third, asset validation checks that all objects are not in the forbidden set and, for containment tasks (e.g., ``place $b_i$ inside $b_j$''), that inner objects fit inside containers with some clearance.
Fourth, if validation fails, feedback is gathered into a fix prompt $\mathcal{Q}_{\text{fix}}$ that includes the original prompt $\mathcal{Q}$, the invalid output, and an error message $\mathcal{E}$ describing syntax errors or invalid asset references.
The fix prompt is provided to the LLM to refine the task and repeat the process.

\subsection{Benchmark Design} \label{sec:method:benchmark}
Inspired by Visual Question and Answering (VQA) benchmarks, our benchmark design enables enables evaluating specific competency axes:
\begin{itemize}[noitemsep, topsep=0pt, parsep=0pt, partopsep=0pt,label={},leftmargin=0pt, itemindent=0pt]
\item \textbf{Visual Competency:} Assesses recognition of \emph{color}, \emph{semantics}, and \emph{size}, capturing the policy’s capability to link perceptual attributes with higher-level reasoning.
\item \textbf{Procedural Competency:} Evaluates the ability to perform tasks that involve action-oriented reasoning, including \emph{affordances}, \emph{reorientation}, or \emph{stacking}.
\item \textbf{Relational Competency:} Tests understanding of language \emph{conjunctions} (e.g., `and', `or'), \emph{counting}, and \emph{spatial} relationships, measuring how effectively the policy interprets multi-object instructions and scene structure.
\end{itemize}
Figure~\ref{fig:rqas} shows examples of these questions accompanied by scene examples.
Since any one task cannot be categorized as containing exactly evaluating one attribute, each task is labeled with one or more attributes belonging to one or more competencies.

In RoboLab, each task can be decomposed into a \emph{sequential list of subtasks, each containing parallel events}.
For example, the task \emph{``Put the apple and orange on the plate, then put the banana in the bowl''} decomposes to subtasks \texttt{PickPlace(orange) \& PickPlace(apple)} followed by subtask \texttt{PickPlace(banana)}, where each \texttt{PickPlace} contains parallel events (Grasp $\rightarrow$ Hover $\rightarrow$ Drop $\rightarrow$ Done) for each object.

Difficulty of tasks in our benchmarking system is rated using the following formula, based on the task length and require level of competency:
$\mathrm{Difficulty Score} = N_\mathrm{subtasks} + \max(w_\mathrm{skill})$
where $w_\mathrm{skill}$ is 0 for visual identification, 1 for spatial reasoning, 2 for procedural reasoning, and 3 for reorientation and dynamic tasks.
We use this system based on how much reasoning and dexterity should be required in order for the task to be complete.
Based on the difficulty score, tasks fall into one the following difficulty levels: \textit{simple} ($\leq 2$), \textit{moderate} ($3$--$4$), or \textit{complex} ($\geq 5$). The difficulty levels allows the user to quickly analyze the results at a glance, although we provide a more granular analysis is possible by examining scene composition and task horizon, discussed in Sec.~\ref{sec:results:all}.

\subsection{Metrics for Evaluation} \label{sec:metrics}

While task success rate remains a fundamental metric, prior work~\cite{kressgazit2024bestpractices} has demonstrated that they fail to reveal nuanced aspects of policy behavior and failure modes. Unlike approaches relying on human judgment~\cite{robotarena}, we define a set of discrete and continuous metrics to characterize policy performance. All of the following measures are independent measures and together paint a complete picture of policy bias.

\paragraph{Normalized Scores}
We compute a \emph{normalized graded score} for each task based on the subtasks $\tau$ as follows: \(Sc(\mathcal{T}) = \frac{1}{|\mathcal{T}|} \sum_{\tau \in \mathcal{T}} w_{\tau} Sc(\tau)\), where $Sc(\tau) = ||\sum_{e} w_e||$ is the subtask score given events $e$. By default, all events and subtask weights are equal. Subtask/event weights default to equal but are user-configurable per task, allowing consequential events (e.g., an initial grasp in a cluttered scene) to be weighted appropriately. Successful episodes will have a score of 1.

\paragraph{Language Variations}
Tasks in \framework are paired with a set of language instructions that the evaluator can query from. Having a set of high-variance language instructions is crucial for task-generalist policies: Given language variants of the same underlying task, the policy should behave similarly. This provides insight into the reasoning failures.

\paragraph{Trajectory Metrics}
Trajectory quality metrics capture characteristics of motion efficiency and optimality. We compute the following:
\emph{Spectral arc-length (SPARC)}, which evaluates motion smoothness~\cite{sparc} via the arc length of the normalized Fourier magnitude spectrum of the velocity profile. Given a speed profile $v(t)$ of the end effector over time interval $[0, T]$
\begin{equation}
\text{SPARC} = -\int_0^{\omega_c} \sqrt{\left(\frac{1}{\omega_c}\right)^2 + \left(\frac{d\hat{V}(\omega)}{d\omega}\right)^2} \, d\omega
\end{equation}
where $\hat{V}(\omega) = V(\omega)/V(0)$ represents the normalized Fourier magnitude spectrum.
Smoother motions yield values closer to zero, while jerkier trajectories produce more negative values.
We employ an adaptive cutoff frequency $\omega_c = \min(10\text{ Hz}, \omega_{\alpha})$, where $\omega_{\alpha} = \max_{k \in \mathcal{K}} \omega_k$ and $\mathcal{K} = \{ k \mid \hat{V}(\omega_k) \geq \alpha \}$ denotes the set of frequency bins exceeding threshold $\alpha = 0.05$.
This adaptive strategy ensures that the smoothness evaluation focuses on relevant frequency components.
Lastly, trajectory optimality is assessed through end effector \emph{speed} $v(t)$,  and \emph{path length} $l = \sum_{k=0}^{N-1} \| p_{k+1} - p_k \|$, where $p_k$ denotes the end-effector position at timestep $k$.
Shorter path lengths indicate more direct trajectories and generally reflect superior motion quality.

\begin{table*}[bth]
    \centering
    \caption{\small \textbf{Overall performance of SOTA policies on RoboLab.} While recent foundation models exhibit emerging capabilities across diverse task dimensions, overall performance remain limited.
    See Table~\ref{table:categories_details} and Table~\ref{table:categories_details_traj} for more granular breakdown of performance trajectory-quality results.
    }
    \label{table:categories_summary}

    \resizebox{\textwidth}{!}{%
    \begin{tabular}{l||ccc||ccc||ccc|ccc|ccc}
    \toprule
    & \multicolumn{3}{c||}{\cellcolor{gray!15}\textbf{Overall Metrics}} & \multicolumn{3}{c||}{\cellcolor{gray!15}\textbf{Difficulty} (succ\% / score)} & \multicolumn{3}{c|}{\cellcolor{gray!15}\textbf{Procedural} (succ\% / score)} & \multicolumn{3}{c|}{\cellcolor{gray!15}\textbf{Relational} (succ\% / score)} & \multicolumn{3}{c}{\cellcolor{gray!15}\textbf{Visual} (succ\% / score)} \\
    \rowcolor{nvidiagreen!15}
    \textbf{Model} & Succ\% / Score ($\uparrow$) & SPARC ($\uparrow$) & Speed ($\uparrow$) & \scriptsize{simple} & \scriptsize{moderate} & \scriptsize{complex} & \scriptsize{affordance} & \scriptsize{reorientation} & \scriptsize{stacking} & \scriptsize{conjunction} & \scriptsize{counting} & \scriptsize{spatial} & \scriptsize{color} & \scriptsize{semantics} & \scriptsize{size} \\
    \midrule
    \piZeroFive~\cite{pi05:intelligence2025pi05visionlanguageactionmodelopenworld} & \textbf{28.0 / 0.43} & $-$8.34 {\scriptsize\textcolor{gray}{$\pm$6.7}} & \textbf{5.4 {\scriptsize\textcolor{gray}{$\pm$1.6}}} & \textbf{29.7 / 0.39} & \textbf{31.5 / 0.51} & \textbf{13.5 / 0.44} & \textbf{10.0 / 0.27} & \textbf{28.3 / 0.48} & \textbf{35.0 / 0.57} & \textbf{43.8 / 0.55} & \textbf{65.7 / 0.75} & \textbf{21.0 / 0.36} & \textbf{23.8 / 0.42} & \textbf{23.2 / 0.39} & \textbf{30.0 / 0.36} \\
    \piZeroFast~\cite{pi0-fast:pertsch2025fastefficientactiontokenization} & 15.5 / 0.27 & $-$9.63 {\scriptsize\textcolor{gray}{$\pm$5.4}} & 4.6 {\scriptsize\textcolor{gray}{$\pm$1.8}} & 20.2 / 0.27 & 13.3 / 0.29 & 2.9 / 0.22 & 2.5 / 0.13 & 3.3 / 0.10 & 15.0 / 0.28 & 27.5 / 0.38 & 44.3 / 0.59 & 15.9 / 0.30 & 6.2 / 0.17 & 10.5 / 0.22 & 10.0 / 0.18 \\
    \groot~\cite{gr00tn1_2025} & 7.2 / 0.17 & \textbf{$-$6.87 {\scriptsize\textcolor{gray}{$\pm$4.6}}} & 4.3 {\scriptsize\textcolor{gray}{$\pm$1.4}} & 8.8 / 0.14 & 7.9 / 0.25 & 0.0 / 0.12 & 0.8 / 0.11 & 0.0 / 0.03 & 8.3 / 0.18 & 10.0 / 0.14 & 18.6 / 0.34 & 7.2 / 0.22 & 3.8 / 0.16 & 6.8 / 0.17 & 0.0 / 0.03 \\
    \piZero~\cite{pi0:black2026pi0visionlanguageactionflowmodel} & 5.0 / 0.12 & $-$9.49 {\scriptsize\textcolor{gray}{$\pm$4.1}} & 4.2 {\scriptsize\textcolor{gray}{$\pm$1.3}} & 7.2 / 0.10 & 3.6 / 0.18 & 0.0 / 0.09 & 0.0 / 0.07 & 1.7 / 0.06 & 0.0 / 0.05 & 12.5 / 0.21 & 11.4 / 0.28 & 3.1 / 0.16 & 1.2 / 0.08 & 2.3 / 0.09 & 1.7 / 0.03 \\
    \paliGemma~\cite{paligemma:beyer2024paligemmaversatile3bvlm} & 3.4 / 0.10 & $-$16.52 {\scriptsize\textcolor{gray}{$\pm$10.2}} & 1.9 {\scriptsize\textcolor{gray}{$\pm$1.6}} & 3.4 / 0.06 & 4.9 / 0.16 & 0.0 / 0.09 & 0.0 / 0.01 & 1.7 / 0.07 & 0.0 / 0.06 & 3.8 / 0.06 & 22.9 / 0.36 & 1.7 / 0.12 & 0.8 / 0.09 & 3.5 / 0.09 & 0.0 / 0.05 \\
    \bottomrule
    \end{tabular}
    }

\end{table*}

\subsection{Sensitivity Analysis}
We present a Bayesian framework for evaluating policy robustness across diverse environmental conditions using Simulation-Based Inference (SBI).
This analysis provides insight into which scene parameters are most strongly linked to success and failure outcomes by learning an approximate posterior distribution over them given evaluation data.
Let $\theta = (\theta^{\text{cont}}, \theta^{\text{disc}})$ denote the environment parameters comprising of continuous variables (e.g., object distance, camera displacement) and/or discrete variables.
After evaluating policy $\pi$ under varied conditions, we generate episodes $\mathcal{D} = \{(\theta_i, x_i)\}_{i=1}^{N}$ with observed outcomes $x_i$ (e.g., task success).
The posterior distribution $p(\theta \mid x) \propto p(x \mid \theta) p(\theta)$ is approximated using Mixed Neural Posterior Estimation (MNPE), which trains a neural density estimator $q_\phi(\theta \mid x)$ to directly learn the mapping from observations to parameter distributions.
The resulting posterior $q_\phi(\theta \mid x)$ characterizes which scene variables are most associated with a target observation $x$.
Our approach provides systematic assessment of which variables most strongly influence performance outcomes.
Further details are in Appendix~\ref{app:mnpe}.

\paragraph{Task Adherence}
While task success is determined by the target world state, the policy may exhibit undesirable behaviors during rollout that do not affect the final outcome.
For example, Fig.~\ref{fig:task-progression} demonstrates a successful episode in which the policy incorrectly grasped an extraneous object. Such errors highlight potential biases in the policy not captured by other metrics.
Thus, capturing discrete events such as grasping wrong objects, executing redundant actions and unintended collisions,  highlights reasoning and task adherence failures.
Our benchmark automatically records instances of events; including wrong object grasped, object dropped, and gripper collisions.

\begin{figure}[b!]
    \centering
    \includegraphics[width=1\linewidth]{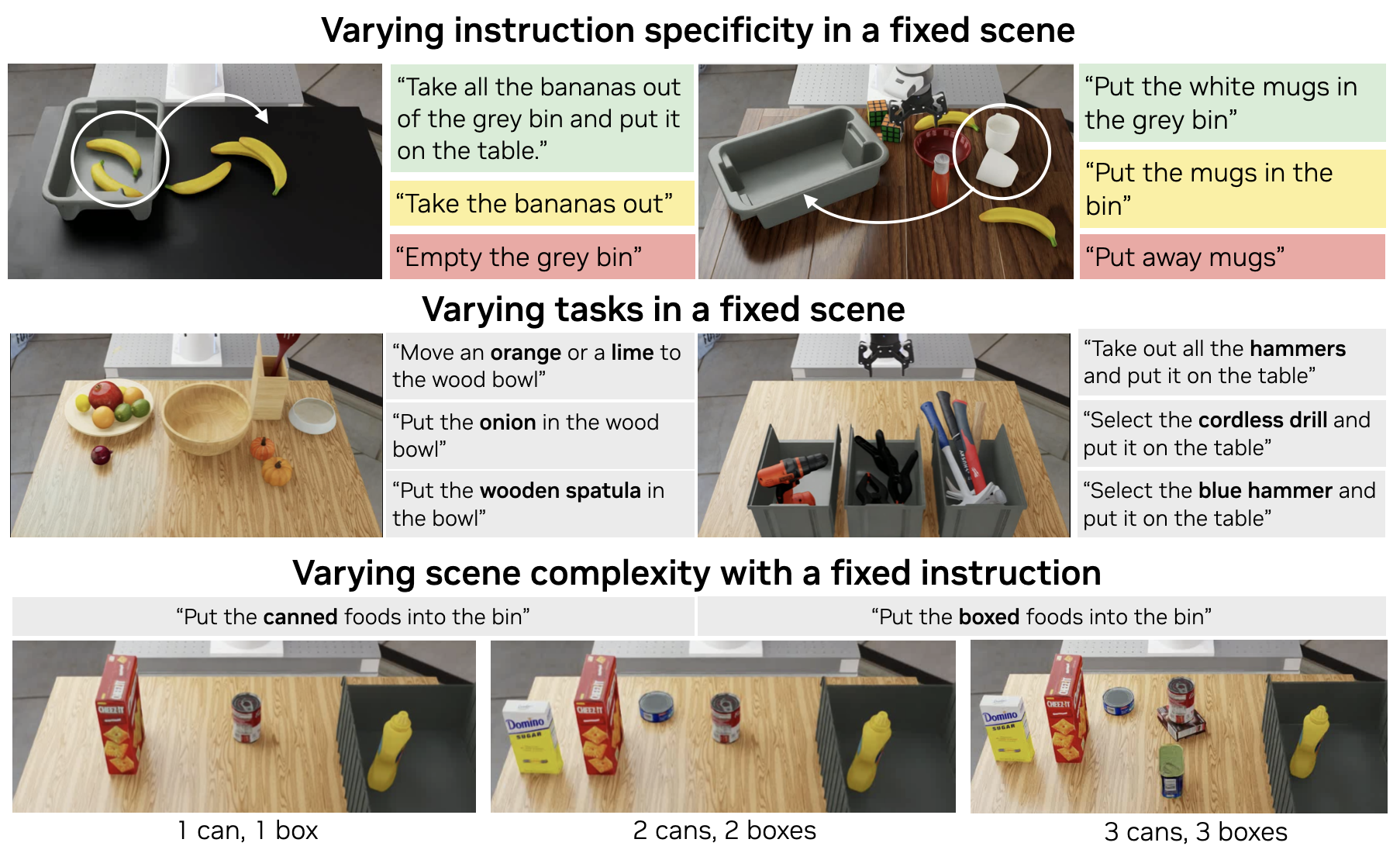}
    \caption{\small Examples of language ablation experiments. Top: Same scene and goal, but the instruction wording ranges from precise to increasingly vague. Middle: Same scene, but the instruction specifies different tasks to perform. Bottom: Same instruction, but the scene becomes progressively more complex.
    }
    \label{fig:language-tasks}
\end{figure}

\begin{figure*}[t]
    \centering
    \begin{subfigure}{0.32\linewidth}
        \centering
        \includegraphics[width=\linewidth]{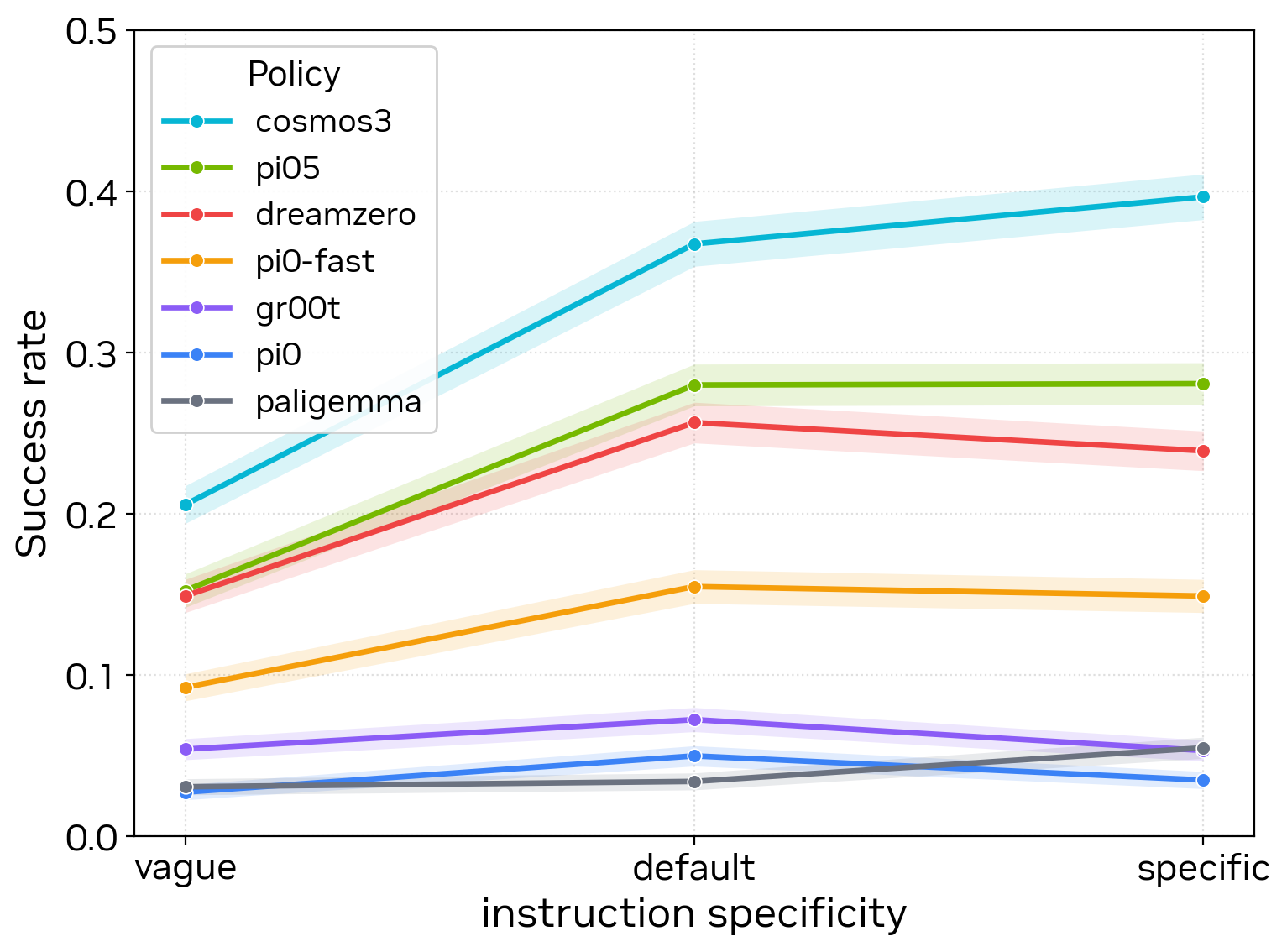}
        \caption{\small SR relative to instruction specificity}
        \label{fig:succ_by_specificity}
    \end{subfigure}
    \hfill
    \begin{subfigure}{0.32\linewidth}
        \centering
        \includegraphics[width=\linewidth]{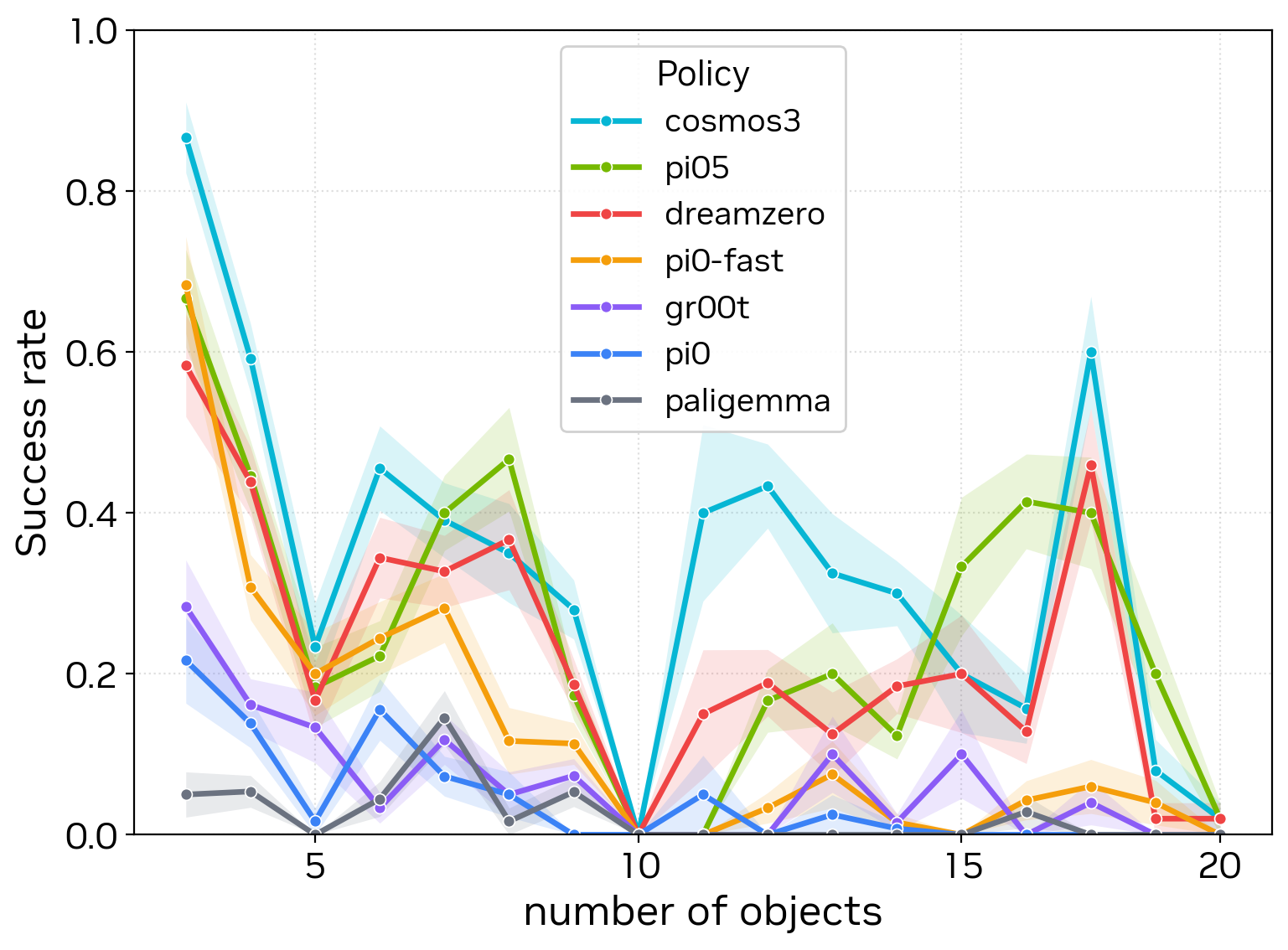}
        \caption{\small SR relative to scene complexity}
        \label{fig:succ_by_num_objects}
    \end{subfigure}
    \hfill
    \begin{subfigure}{0.32\linewidth}
        \centering
        \includegraphics[width=\linewidth]{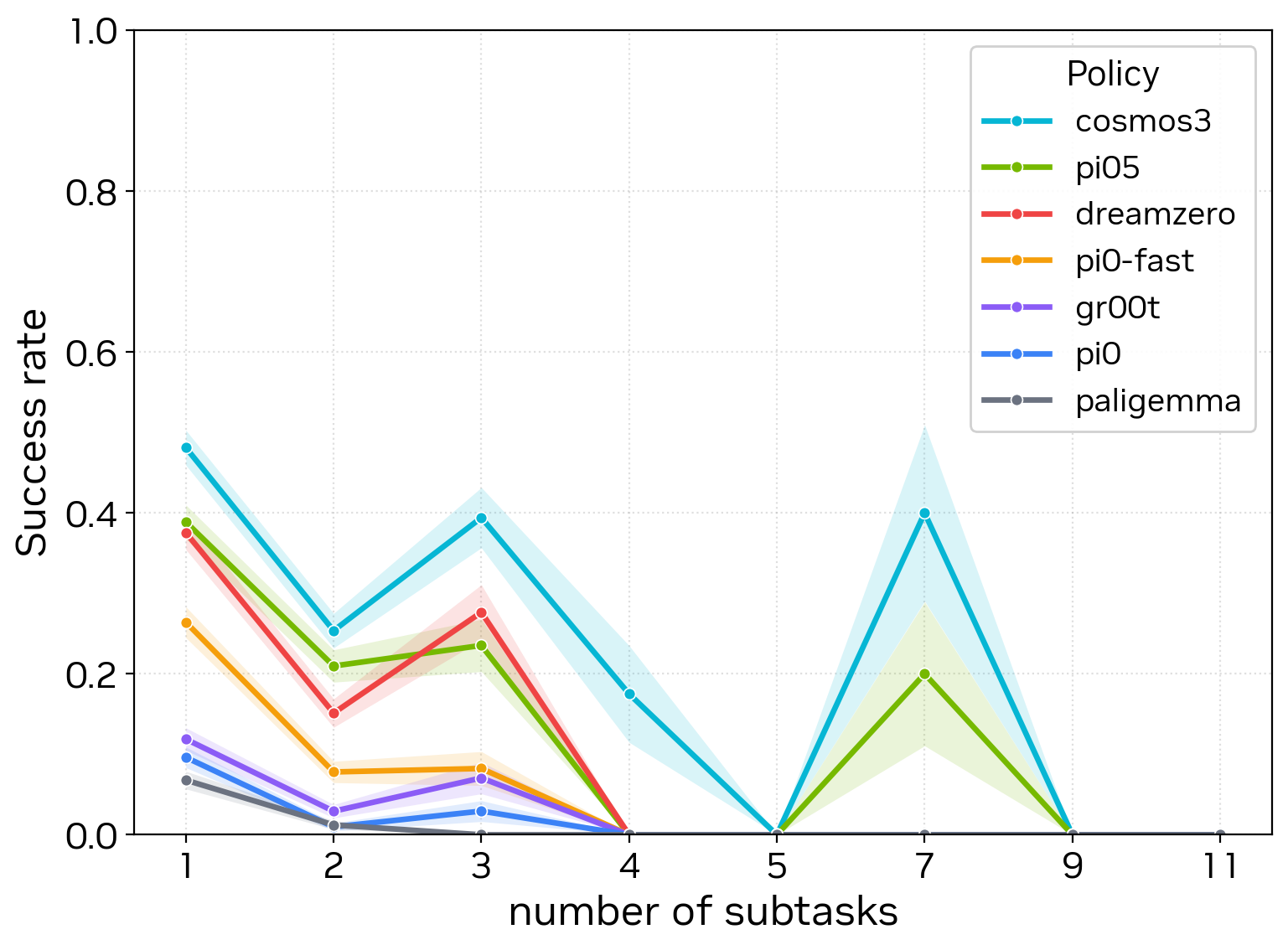}
        \caption{\small SR relative to task horizon}
        \label{fig:succ_by_subtasks}
    \end{subfigure}
    \caption{\small \textbf{Effect of language, scene complexity, and task horizon on policy performance (success rate).} Performance using success rate across complexity axes. For a robust policy, performance should relatively comparable as complexity increases. (a) Performance degrades as instructions become more abstract, indicating brittle task-level reasoning. (b) Performance degrades sharply as the number of objects in the scene increases, exposing systematic geometric biases in current models. (c) Performance degrades as the task horizon increases, indicating limited multi-step reasoning across all evaluated policies. Fig.~\ref{fig:score_complexity} shows the same analysis with normalized scores for a more granular analysis.}
    \label{fig:succ_complexity}
\end{figure*}

\section{Experiments}

We evaluate several off-the-shelf generalist robotics models, performing controlled ablations, and environmental perturbations to observe where failures concentrate.
The experiments are designed to address the following questions:
\textbf{Q1:} How well does today's SOTA models generalize, given varying language instructions, scene complexity, and environment pertubations?
\textbf{Q2:} When and why does a policy fail?
\textbf{Q3:} Can a simulated benchmark be used to evaluate real-world models?

\subsection{Experiment Setup}
We evaluate SOTA models on \benchname, a benchmark containing 120 tasks of varying difficulty levels (65 simple, 38 moderate, 18 complex).
We evaluate policies finetuned on the DROID robot~\cite{droid}, which is an open-sourced robot used for VLAs~\cite{polaris, roboarena}.
DROID has a 7-DOF Franka Panda robot arm with a Robotiq-2F-85 gripper, an externally mounted ZED 2i camera with $f$$=$$2.1$mm, and a ZED mini as the wrist camera.
All policies were fine-tuned on the DROID dataset~\cite{droid}: \piZeroFive~\cite{pi05:intelligence2025pi05visionlanguageactionmodelopenworld}, \piZeroFast~\cite{pi0-fast:pertsch2025fastefficientactiontokenization}, \piZero~\cite{pi0:black2026pi0visionlanguageactionflowmodel}, \paliGemma~\cite{paligemma:beyer2024paligemmaversatile3bvlm}, \groot~\cite{gr00tn1_2025}.

The action space is 7-DOF Franka joint positions and a 1-DOF binary gripper command.
The environments are composed of a default office-like background and natural lighting to mimic typical setups in the DROID dataset~\cite{droid}, with wrist and external camera poses designed to strongly match the real-world DROID robot.
Each task was repeated with $N$=10 episodes per task. Notes on statistical analysis of $N$ is discussed in Appendix~\ref{sec:stats}.

\subsection{Task Results} \label{sec:results:all}

Table~\ref{table:categories_summary} shows the overall results on \benchname, highlighting success rate and score.
Note that success rates indicates number of episodes that were completed, whereas the score indicates whether if the policy was able to make progress on any task.
Overall success were low, which matches prior observations on out-of-domain generalization for generalist policies~\cite{liberopro}.
More interestingly, the score/success gap reveals capabilities that raw success rates obscure. \piZeroFive{} achieves only 13.5\% success on complex tasks yet attains a score of 0.44, indicating that nearly half of the partial-credit milestones are reached even when full task completion is rare.
This indicates that contemporary policies often partially understand the task but fail in the final stages of execution.

Competency axes also highlighted asymmetric generalization in \emph{relational} reasoning tasks. Overall, policies handled conjunctions and counting better than spatial relations.
In \emph{visual} grounding, performance remained low across attribute types, indicating brittle language-to-object binding beyond a narrow set of familiar object descriptions.
\emph{Procedural} understanding proved most challenging.

To further probe generalization robustness of the policies, we performed analysis on variations on instructions, scene complexity, and task difficulty. An example of these experiments is shown in Fig.~\ref{fig:language-tasks}.

\paragraph{Performance relative to instruction specificity}
Fig.~\ref{fig:succ_by_specificity} reports overall \benchname success rates under three levels of instruction specificity (vague, default, specific) for the same underlying tasks and scenes.
We observe degradation as instructions become more abstract (e.g., \piZeroFive drops from 28.0\% on default to 15.3\% on vague)
This sensitivity reveals that current models lack robust reasoning against the task goal.
A truly generalist policy should be invariant to instruction phrasing. A policy that has understood the task should perform comparably whether the instruction is vague or specific.

\paragraph{Performance relative to scene complexity}
Fig.~\ref{fig:succ_by_num_objects} isolates the effect of scene complexity by increasing the number of objects visually present on the table, representing visual clutter.
Success rates degrade as the number of objects in the scene increases for most policies; however, for \piZeroFive, some tasks were able to be achieved even with high visual clutter.
These results indicates that current policies can perform some visual grounding in cluttered scenes; however, for some policies, additional distractors overwhelm the policy's ability to identify and manipulate the correct target.

\paragraph{Performance relative to task horizon}
Fig.~\ref{fig:succ_by_subtasks} shows that performance degrades as the task horizon increases.
Performance degrades as the task horizon increases.
This indicates that current models lack the multi-step reasoning and compositional planning required for longer-horizon tasks.
A performance increase is observed at subtask=7 for \piZeroFive on \texttt{\small CubesAndBlocksInBinTask} is further discussed in Appendix~\ref{sec:subtask7-anomaly}.

Together, these results show how \framework isolates where generalization fails, addressing \textbf{Q1} supporting analysis that can inform data collection and model improvement priorities.

\subsection{Sensitivity Analysis}
We perform a set of variations given two basic tasks and observe the outcome, as illustrated in Fig.~\ref{fig:variations}. These are: 1) Wrist and external camera poses; 2) object poses; 3) Visual features, including background and table textures; and 4) Lighting, including saturation and hue.
Table~\ref{table:variations} illustrates the results for all experiments.

\paragraph{Visual and Lighting variations}
We vary the lighting conditions via color temperature shifts, lighting exposure and strong directional light that generates shadows as the robot is moving.
\textbf{Lighting:} Models were robust to changes in lighting conditions, with 90--100\% success across shadow variations, color temperature shifts, and 500$\times$ intensity changes.
\textbf{Visual appearance:} Variations over 10 background textures and 4 table textures had minimal impact ($<$5\% degradation), suggesting generalization to scene appearance changes.

\begin{figure*}
    \centering
    \includegraphics[width=\linewidth]{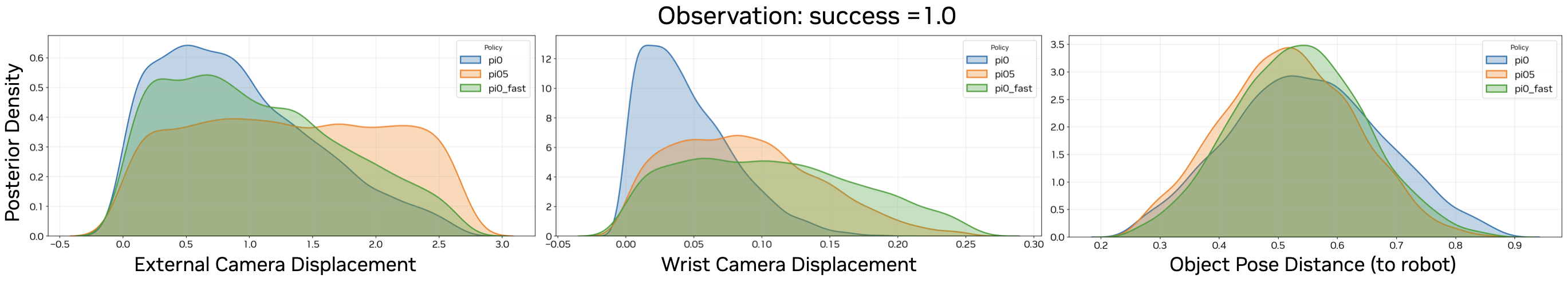}
    \caption{\small Results of the sensitivity analysis using MNPE. Policies were highly sensitive to wrist-camera displacement from the nominal pose, indicating strong dependence on wrist-mounted camera calibration. Success also peaked for objects placed at approximately 0.5m from the robot, likely due to robot reachability.}
    \label{fig:posterior_distribution}
\end{figure*}

\paragraph{Camera variation sensitivity analysis}
We infer posteriors over camera displacement conditioned on task success (Fig.~\ref{fig:posterior_distribution}).
Camera poses were randomized in both orientation and position for 10 episodes each.
Displacement is calculated with respect to the nominal position of the cameras.
Across all policies, the wrist-camera posterior is sharply concentrated near zero, indicating that successful execution often required the wrist camera to remain close to its nominal pose, while performance is more tolerant to external camera position changes.
This indicates performance is critically dependent on wrist camera than external camera.

\paragraph{Object pose variation sensitivity analysis}
We randomize initial object poses via a uniform distribution of 10cm, 20cm, and 30cm within its nominal placement (in front of the robot).
We then infer posteriors over initial object poses conditioned on task success (see Fig.~\ref{fig:posterior_distribution}), relative to the robot pose.
We observe a strong peak over 0.5m from the robot's origin, suggesting that objects placed at this distance has the highest probability of success, likely due to reachability.

Together, these results show how \framework's analysis framework can be used to systematically identify factors contributed to policy failure, addressing \textbf{Q2}.

\subsection{Real-World Verification}
\begin{figure}
    \centering
    \includegraphics[width=0.8\linewidth]{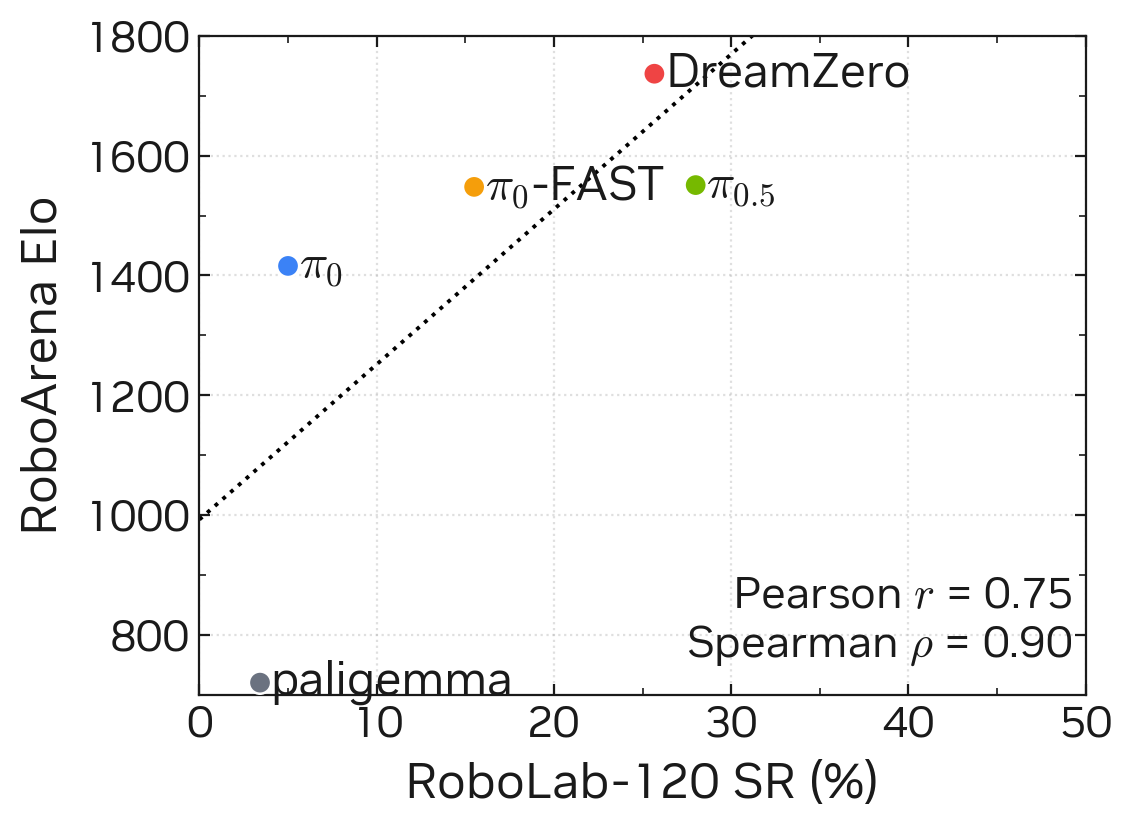}
    \caption{\small Correlation between \benchname{} success rate and RoboArena Elo score across the four policies with both measurements available (\piZeroFive, \piZeroFast, \piZero, \paliGemma). Rankings are preserved (Spearman $\rho = 1.00$) and the scores are positively correlated (Pearson $r = 0.68$), suggesting \benchname{} performance is a useful proxy for real-world policy quality.}
    \label{fig:roboarena_correlation}
\end{figure}

To assess whether \framework can be used as a proxy for real-world evaluation, we compare performance results from \framework against RoboArena~\cite{roboarena}, an open-source real-world benchmarking system.
Fig.~\ref{fig:roboarena_correlation} plots per-policy \benchname success rate against RoboArena Elo scores.
We observe that the ranking between policies is strongly preserved (Spearman $\rho = 1.00$).
Because RoboArena reports Elo while RoboLab reports success rates, Spearman rank correlation is more useful as a proxy, since it measures whether the two benchmarks induce the same ordering over policies.
This indicates that \framework achieves benchmark-level correlation with real-world performance, addressing \textbf{Q3}.
We leave deeper investigation of task-level and motion-level correlation to future work.

\begin{table}
    \centering
    \caption{\small Robustness to controlled environmental variations over two simple tasks (\texttt{\small BananaInBowl}, \texttt{\small BananaAndCubeInBowl}). \paliGemma is excluded as it fails to achieve meaningful results.}
    \label{table:variations}
    \resizebox{\linewidth}{!}{%
    \footnotesize
    \begin{tabular}{lrrrrrr}
    \toprule
    & \multicolumn{2}{c}{\textbf{\piZeroFive}} & \multicolumn{2}{c}{\textbf{\piZeroFast}} & \multicolumn{2}{c}{\textbf{\piZero}} \\
    \cmidrule(lr){2-3} \cmidrule(lr){4-5} \cmidrule(lr){6-7}
    \rowcolor{nvidiagreen!15}\textbf{Variation} & Succ.\% & Time (s) & Succ.\% & Time (s) & Succ.\% & Time (s) \\
    \midrule
    \rowcolor{gray!15}\multicolumn{7}{l}{\textit{Lighting}} \\ \midrule
    Color & 96.7 & 14.5 {\scriptsize\textcolor{gray}{$\pm$7.9}} & 93.3 & 17.9 {\scriptsize\textcolor{gray}{$\pm$10.7}} & 6.7 & 31.1 {\scriptsize\textcolor{gray}{$\pm$4.3}} \\
    Shadows & 100.0 & 16.0 {\scriptsize\textcolor{gray}{$\pm$6.0}} & 90.0 & 12.4 {\scriptsize\textcolor{gray}{$\pm$3.5}} & 0.0 & - \\
    Dim & 90.0 & 9.1 {\scriptsize\textcolor{gray}{$\pm$2.1}} & 70.0 & 13.1 {\scriptsize\textcolor{gray}{$\pm$2.8}} & 70.0 & 35.5 {\scriptsize\textcolor{gray}{$\pm$9.7}} \\
    Overexposed & 100.0 & 13.9 {\scriptsize\textcolor{gray}{$\pm$4.4}} & 100.0 & 9.6 {\scriptsize\textcolor{gray}{$\pm$1.7}} & 0.0 & - \\ \midrule
    \rowcolor{gray!15}\multicolumn{7}{l}{\textit{Visual Variations}} \\ \midrule
    Background & 85.0 & 14.4 {\scriptsize\textcolor{gray}{$\pm$8.7}} & 70.0 & 21.3 {\scriptsize\textcolor{gray}{$\pm$10.7}} & 25.0 & 31.6 {\scriptsize\textcolor{gray}{$\pm$11.8}} \\
    Table texture & 87.5 & 19.0 {\scriptsize\textcolor{gray}{$\pm$13.8}} & 60.0 & 19.0 {\scriptsize\textcolor{gray}{$\pm$12.9}} & 22.5 & 28.1 {\scriptsize\textcolor{gray}{$\pm$6.9}} \\ \midrule
    \rowcolor{gray!15}\multicolumn{7}{l}{\textit{Object Pose}} \\ \midrule
    10cm & 95.0 & 16.2 {\scriptsize\textcolor{gray}{$\pm$9.2}} & 55.0 & 26.5 {\scriptsize\textcolor{gray}{$\pm$13.8}} & 22.5 & 34.7 {\scriptsize\textcolor{gray}{$\pm$13.3}} \\
    20cm & 95.0 & 19.7 {\scriptsize\textcolor{gray}{$\pm$10.2}} & 40.0 & 21.8 {\scriptsize\textcolor{gray}{$\pm$9.5}} & 20.0 & 37.4 {\scriptsize\textcolor{gray}{$\pm$11.2}} \\
    30cm & 62.5 & 18.9 {\scriptsize\textcolor{gray}{$\pm$8.7}} & 35.0 & 24.3 {\scriptsize\textcolor{gray}{$\pm$11.9}} & 17.5 & 24.3 {\scriptsize\textcolor{gray}{$\pm$11.9}} \\ \midrule
    
    \rowcolor{gray!15}\multicolumn{7}{l}{\textit{Camera Pose}} \\ \midrule
    external & 85.0 & 17.4 {\scriptsize\textcolor{gray}{$\pm$11.3}} & 45.0 & 27.7 {\scriptsize\textcolor{gray}{$\pm$10.6}} & 50.0 & 27.4 {\scriptsize\textcolor{gray}{$\pm$16.0}} \\
    wrist & 60.0 & 21.9 {\scriptsize\textcolor{gray}{$\pm$13.4}} & 25.0 & 20.1 {\scriptsize\textcolor{gray}{$\pm$9.9}} & 10.0 & 35.3 {\scriptsize\textcolor{gray}{$\pm$10.4}} \\ %
    \bottomrule
    \end{tabular}
    }
    
    \end{table}

\section{Limitations}

While \framework provides a flexible and scalable framework for evaluating language-conditioned manipulation, it currently focuses on rigid-body tabletop scenes and does not fully capture the challenges of deformable object manipulation (e.g., cloth, cables, bags).
Moreover, many contact-rich skills that require precise force control, compliant interaction, or complex frictional dynamics are underrepresented and dependent on the physics simulation fidelity, limiting Robolab’s coverage of fine-grained, low-level control tasks.
Our subtask evaluation system breaks down for open-ended and ambiguous long-horizon tasks (e.g., ``clean up all the laundry'') and human judgment becomes necessary.
However, given that current policies achieve fairly low scores, this is not yet a practical bottleneck, and our approach remains valid.
Finally, although evaluation in high-fidelity simulation is a strong proxy for real-world performance, a residual visual distribution shift remains. This gap needs to be characterized further both by analyzing the behavior and robustness of the visual perception stack and through extensive validation on real-world deployments.

\section{Conclusion}

Recent benchmarking efforts have made significant strides in scalable robot evaluation, but they primarily assess robustness to perturbations of training environments rather than true task generalization to novel scenarios.
\framework addresses this gap by evaluating real world policies in a high-fidelity simulation, structured evaluation vectors that decompose policy competence into visual, procedural, and relational dimensions, and a set of sensitivity analysis set of novel analysis that provides insight into policy behavior for robotics.
Our benchmarking framework enables the community to critically answer the question of \emph{generalization} and \emph{performance}.
At the same time, the framework is designed to be pragmatically usable: new tasks can be authored in minutes by arranging objects on a tabletop and attaching language instructions, and a generative scene–task–environment workflow that supports continuous benchmark evolution.

\bibliography{bib}
\bibliographystyle{plainnat}

\clearpage
\appendices

\section*{Acknowledgements}
We thank Arhan Jain, Karl Pertsch, Alperen Degirmenci, Valts Blukis, and Siyi Chen for helpful discussions in building RoboLab.

\section{Details on the \framework Benchmark} \label{app:benchmark}
In this section we provide detail on the benchmark.

\framework provides a set of $\sim$300 object assets from well-known 3D pose estimation benchmarks; including YCB~\cite{ycb}, HOT3D~\cite{hot3d}, HOPE~\cite{hope}, HANDAL~\cite{handal}, and VoMP~\cite{dagli2025vomp}.
Each asset contains a visual and collision mesh, with mass and friction properties added.
Each object has a language description and an object label attached to it.
This forms the catalog of objects used in the scenes, in either manual or LLM-scaled scene environments.

Along with \framework, we introduce an initial benchmark, \benchname, which contains \numtasks manually generated tasks. Details on the scenes and tasks, including images and language instructions is available along with full benchmark results is available on the open source repository \url{https://github.com/NVlabs/RoboLab/}.

\subsection{Statistical significance of results} \label{sec:stats}
Due to stochastic nature of simulators and the policies, it is recommended to run multiple episodes per task per policy \cite{lbm}.
We evaluate each policy on N=10 episodes per task, and report results on our benchmark as aggregate results instead of per-task analysis between policies.

With only 10 trials, the 95\% confidence interval on a single \emph{per-task} success rate is approximately $\pm$30\% near p=0.5 and $\pm$19\% near p=0.9, meaning per-task numbers should be interpreted as coarse indicators rather than precise estimates.
Aggregate scores across all tasks are considerably tighter, since the effective sample size scales with the number of tasks, but per-task claims and fine-grained policy comparisons remain underpowered.
For results that support per-task conclusions or resolve differences smaller than ~10\% between policies, we recommend increasing N to at least 100 episodes per task, which reduces the 95\% CI half-width to roughly $\pm$10\% at p=0.5 and $\pm$6\% at p=0.9, giving meaningful results for per-task analysis between policies.

\subsection{Comparison against other benchmarks} \label{sec:benchmark-comparison}
We describe a comparison of existing methods for task generation, which we categorize into \emph{high vs low‑overhead} in the following table.

All three demand considerable manual effort per scene, making large-scale scene/task creation tedious.
In contrast, RoboLab lets users compose scenes via drag-and-drop, and tasks via manual text specification or invoke an AI agent to generate N tasks per scene at once.

\begin{table}[htb!]
    \centering
    \caption{Comparison of task creation pipelines across benchmarks. O/H denotes overhead.}
    \label{table:benchmark-overhead}
    {\small
    \begin{tabular}{@{}p{1.3cm}p{5.0cm}p{0.3cm}@{}}
    \toprule
    \textbf{Benchmark} &  $\;$ \textbf{Pipeline (per 1 scene 1 task)} & \textbf{O/H} \\
    \midrule
    RobotArena $\infty$ & VLM+2D bbox detector per frame$\rightarrow$2D‑to‑3D diffusion$\rightarrow$3D asset retrieval$\rightarrow$diff.\ rendering & High \\
    Polaris & RGB-D video$\rightarrow$camera calib.$\rightarrow$3DGS$\rightarrow$mesh extraction, semantic segmentation$\rightarrow$physics & High (1hr) \\
    LIBERO & Object+placement+task+goal (python)$\rightarrow$BDDL & High \\
    \textbf{RoboLab} & \textbf{Drag-and-drop$\rightarrow$text instructions} & \textbf{Low} \\
    \bottomrule
    \end{tabular}
    }
    \end{table}

\begin{figure}[bth]
    \centering
    \begin{subfigure}{0.26\linewidth}
        \centering
        \includegraphics[width=\linewidth]{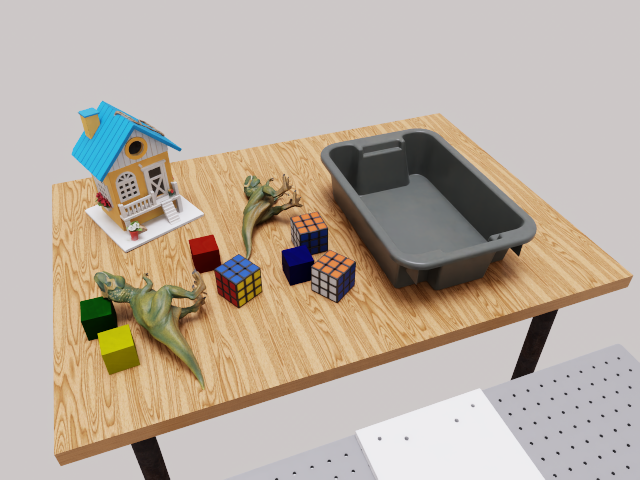}
        \caption{\small Initial scene}
        \label{fig:subtask7_anomaly_initial}
    \end{subfigure}
    \hfill
    \begin{subfigure}{0.35\linewidth}
        \centering
        \includegraphics[width=\linewidth]{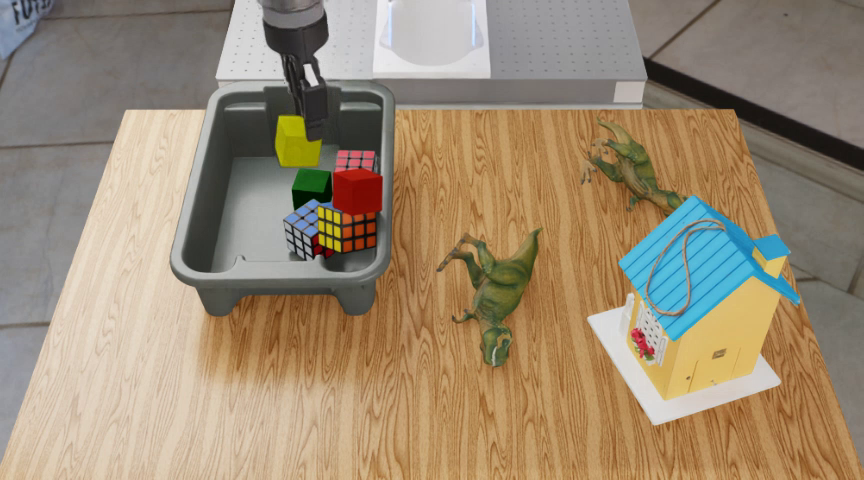}
        \caption{\small succ. eps. 1}
        \label{fig:subtask7_anomaly_env2}
    \end{subfigure}
    \hfill
    \begin{subfigure}{0.35\linewidth}
        \centering
        \includegraphics[width=\linewidth]{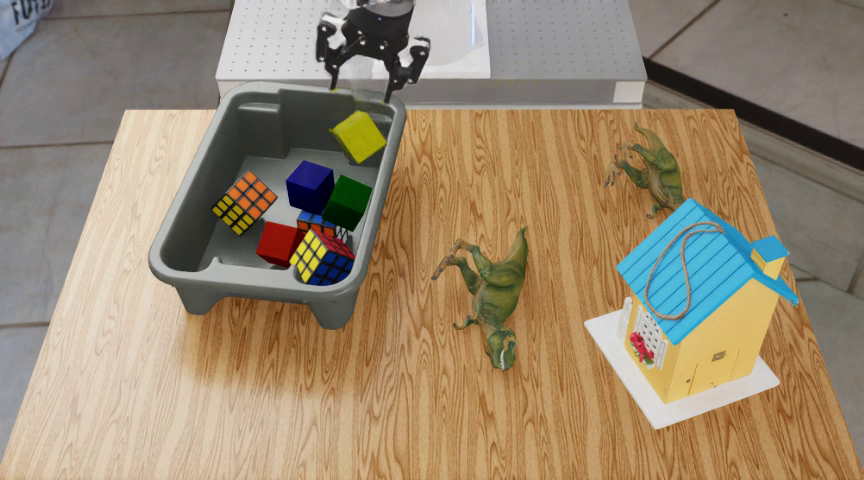}
        \caption{\small succ. eps. 2}
        \label{fig:subtask7_anomaly_env7}
    \end{subfigure}
    \caption{\small Initial scene and final-frame viewports from two successful \piZeroFive{} rollouts on \texttt{\small CubesAndBlocksInBinTask} (7 subtasks). While the task-horizon is long, the task itself is non-causal. The policy was able to make significant progress and completion on this task.}
    \label{fig:subtask7_anomaly}
\end{figure}

\subsection{Score across complexity axes} \label{sec:score-complexity}

\begin{figure*}[t]
    \centering
    \begin{subfigure}{0.32\linewidth}
        \centering
        \includegraphics[width=\linewidth]{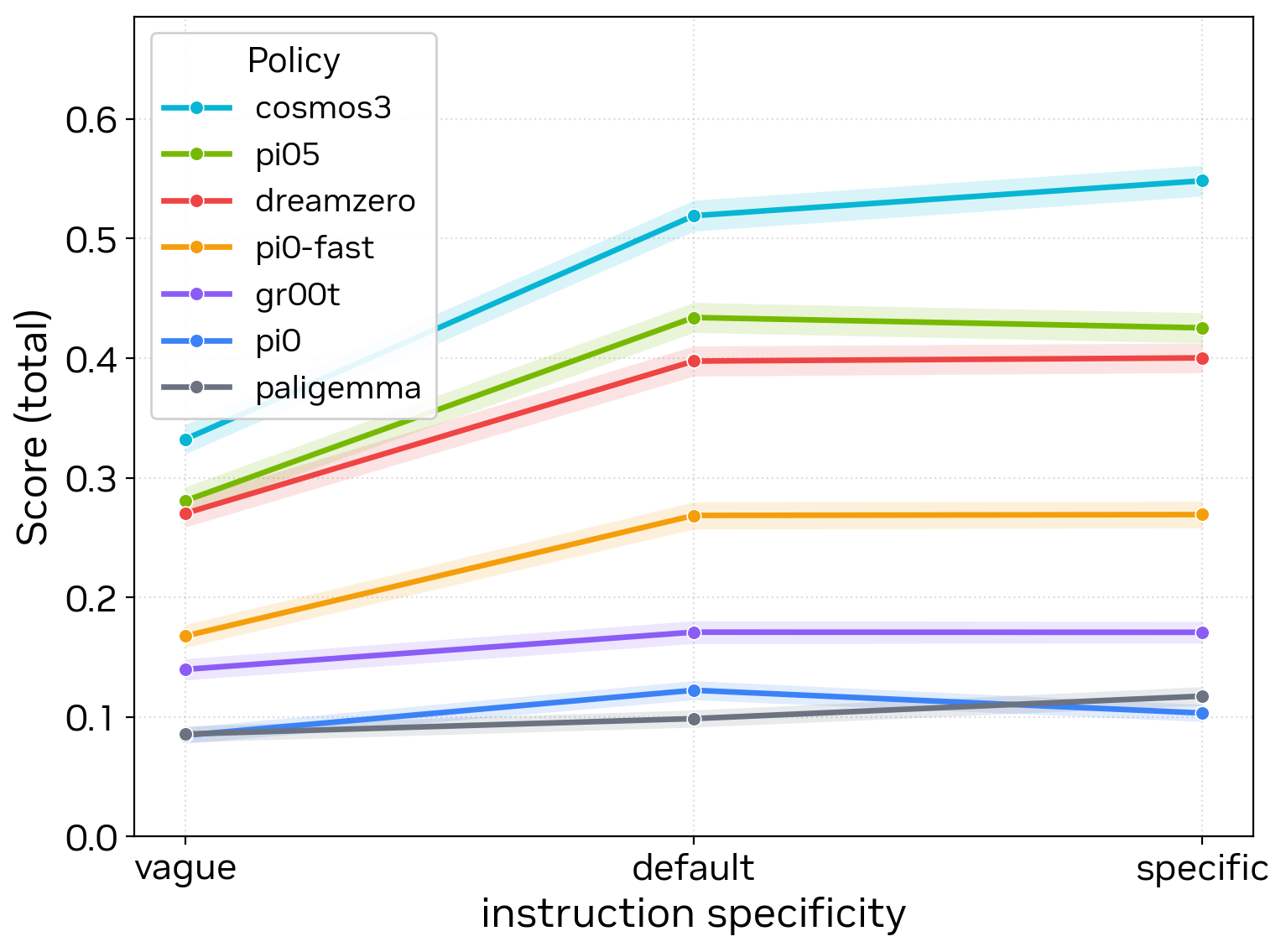}
        \caption{\small Score relative to instruction specificity}
        \label{fig:score_by_specificity}
    \end{subfigure}
    \hfill
    \begin{subfigure}{0.32\linewidth}
        \centering
        \includegraphics[width=\linewidth]{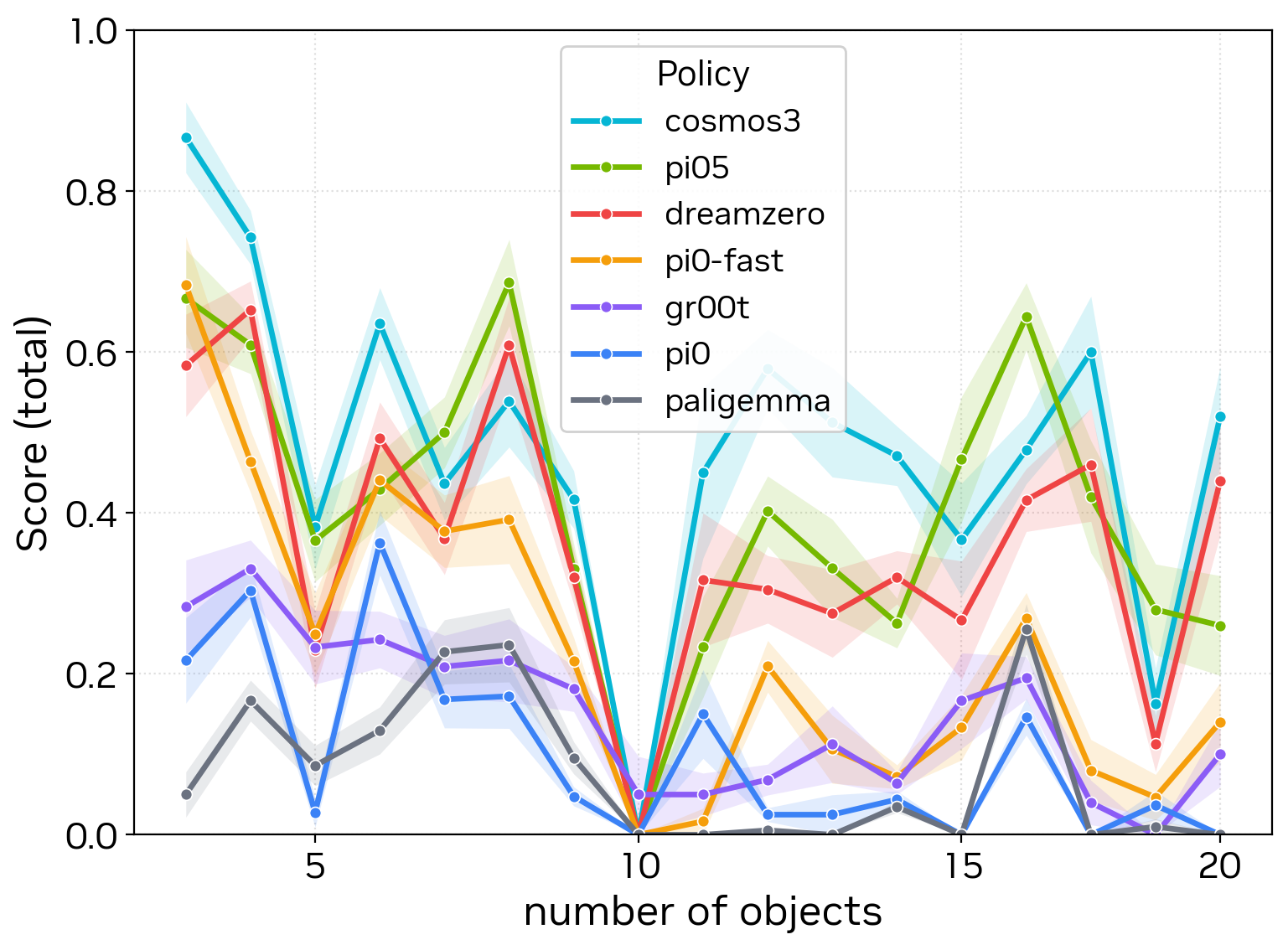}
        \caption{\small Score relative to scene complexity}
        \label{fig:score_by_num_objects}
    \end{subfigure}
    \hfill
    \begin{subfigure}{0.32\linewidth}
        \centering
        \includegraphics[width=\linewidth]{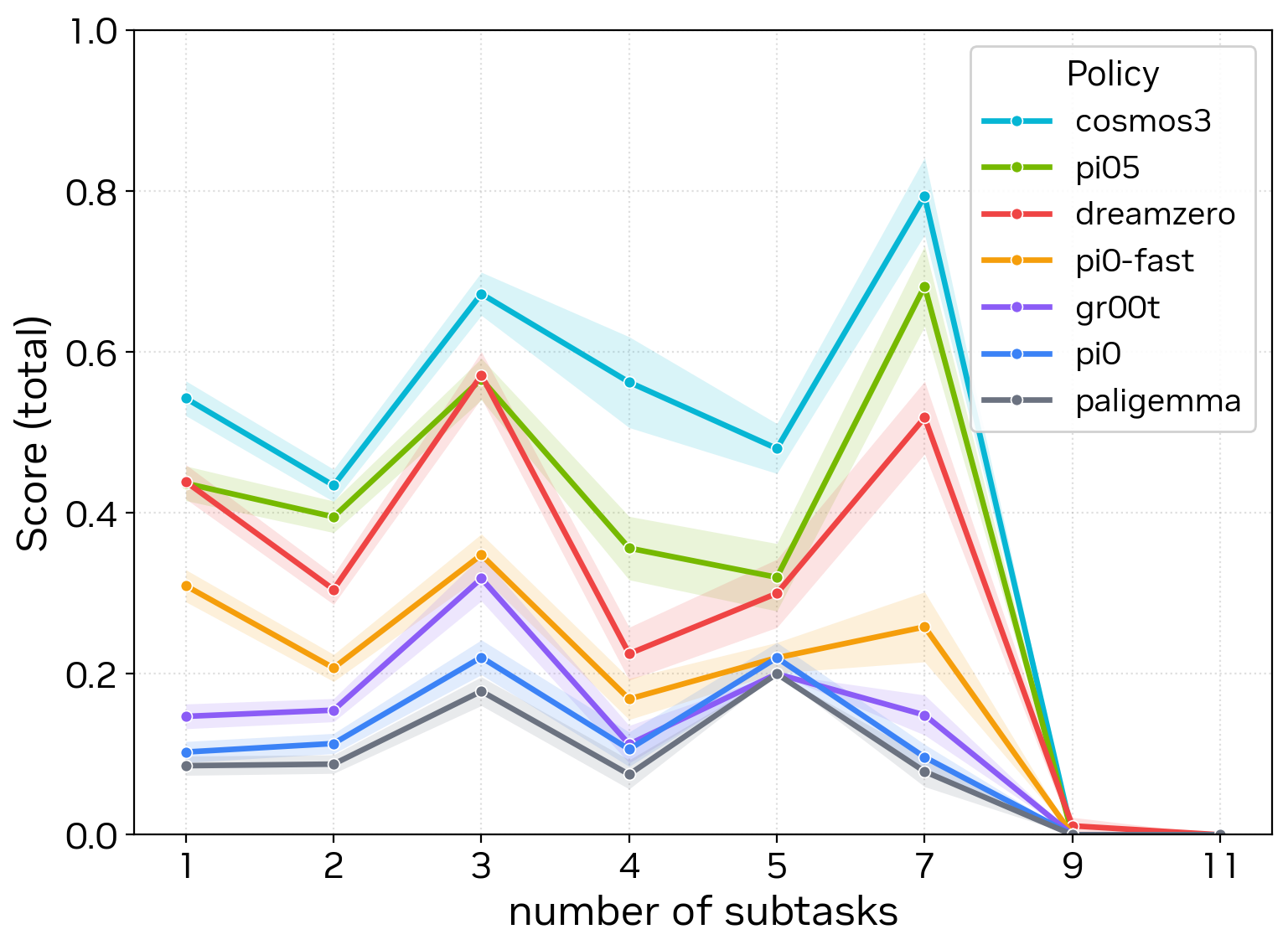}
        \caption{\small Score relative to task horizon}
        \label{fig:score_by_subtasks}
    \end{subfigure}
    \caption{\small \textbf{Score (partial credit) across complexity axes.} Same complexity axes as Fig.~\ref{fig:succ_complexity}, but plotting score (which awards partial credit for failed episodes) instead of success rate. Shaded bands show SE. Score degrades more gracefully than strict success, indicating policies make meaningful progress even when they fail to fully complete the task.}
    \label{fig:score_complexity}
\end{figure*}

For completeness, we illustrate Fig.~\ref{fig:succ_complexity} but with policy score instead of success in Fig.~\ref{fig:score_complexity}.
The gap between the two metrics is largest on \emph{complex} tasks: \piZeroFive{} drops to 13.5\% success but retains a score of 0.44. On complex tasks, \piZeroFive{} still earns 0.35 of the milestones even within failed episodes, indicating that the policy reaches a substantial fraction of subgoals before failing. The same pattern holds in procedural categories where full completion is rare but partial progress is not.

\subsection{Anomalous long-horizon success at subtask=7} \label{sec:subtask7-anomaly}

Fig.~\ref{fig:succ_by_subtasks} shows a non-monotonic spike at $\mathrm{subtask}{=}7$: \piZeroFive{} recovers to $\sim$20\% success and a score of 0.68 at this horizon, despite achieving 0\% at neighboring horizons.
The results show that the $\mathrm{subtask}{=}7$ bin is dominated by \texttt{\small CubesAndBlocksInBinTask} (Fig.~\ref{fig:subtask7_anomaly}).
This task contains repetitive similar pick-and-place sequences with geometrically simple objects (cubes), instead of a true causal long-horizon task where multi-step reasoning is required.
This suggests the apparent ``recovery'' at $\mathrm{subtask}{=}7$ reflects task composition rather than improved long-horizon reasoning. Fig.~\ref{fig:succ_by_subtasks} should therefore be read with this caveat in mind, not as a clean horizon-vs-difficulty curve.

\begin{table*}
    \centering
    \caption{\small \textbf{Capability metrics} per task category on \framework. ``Succ\%'' is the fraction of episodes that fully complete the task; ``Score'' is the mean per-episode normalized score awarded for partial subtask completion (Sec.~\ref{sec:metrics}); ``Score (fail)'' is the mean score of failed episodes only, which isolates partial progress on episodes that did not complete. Trajectory-quality metrics (SPARC, Speed) for the same categories are reported in Table~\ref{table:categories_details_traj}.}
    \label{table:categories_details}
    \resizebox{\textwidth}{!}{%
    \begin{tabular}{lrrrrrrrrrrrrrrrr}
    \toprule
    \rowcolor{nvidiagreen!15} \multirow{2}{*}{\shortstack{\textbf{Task}\\\textbf{Categories}}} & \multirow{2}{*}{\textbf{\#}} & \multicolumn{3}{c}{\textbf{\piZeroFive}} & \multicolumn{3}{c}{\textbf{\piZeroFast}} & \multicolumn{3}{c}{\textbf{\groot}} & \multicolumn{3}{c}{\textbf{\piZero}} & \multicolumn{3}{c}{\textbf{\paliGemma}} \\
    \cmidrule(lr){3-5} \cmidrule(lr){6-8} \cmidrule(lr){9-11} \cmidrule(lr){12-14} \cmidrule(lr){15-17}
     &  & Succ \% & Score & Score (fail) & Succ \% & Score & Score (fail) & Succ \% & Score & Score (fail) & Succ \% & Score & Score (fail) & Succ \% & Score & Score (fail) \\
    \midrule
    \rowcolor{gray!15}Total & 120 & 28.0\% & 0.43 & 0.21 & 15.5\% & 0.27 & 0.13 & 7.2\% & 0.17 & 0.11 & 5.0\% & 0.12 & 0.08 & 3.4\% & 0.10 & 0.07 \\
    \quad simple & 64 & 29.7\% & 0.39 & 0.13 & 20.2\% & 0.27 & 0.08 & 8.8\% & 0.14 & 0.05 & 7.2\% & 0.10 & 0.03 & 3.4\% & 0.06 & 0.03 \\
    \quad moderate & 39 & 31.5\% & 0.51 & 0.28 & 13.3\% & 0.29 & 0.18 & 7.9\% & 0.25 & 0.19 & 3.6\% & 0.18 & 0.15 & 4.9\% & 0.16 & 0.12 \\
    \quad complex & 17 & 13.5\% & 0.44 & 0.35 & 2.9\% & 0.22 & 0.20 & 0.0\% & 0.12 & 0.12 & 0.0\% & 0.09 & 0.09 & 0.0\% & 0.09 & 0.09 \\
    \midrule
    \rowcolor{gray!15}Procedural & 22 & 21.8\% & 0.42 & 0.25 & 6.4\% & 0.16 & 0.11 & 2.7\% & 0.11 & 0.09 & 0.5\% & 0.07 & 0.06 & 0.5\% & 0.04 & 0.04 \\
    \quad affordance & 12 & 10.0\% & 0.27 & 0.19 & 2.5\% & 0.13 & 0.10 & 0.8\% & 0.11 & 0.10 & 0.0\% & 0.07 & 0.07 & 0.0\% & 0.01 & 0.01 \\
    \quad reorientation & 6 & 28.3\% & 0.48 & 0.27 & 3.3\% & 0.10 & 0.07 & 0.0\% & 0.03 & 0.03 & 1.7\% & 0.06 & 0.04 & 1.7\% & 0.07 & 0.05 \\
    \quad stacking & 6 & 35.0\% & 0.57 & 0.35 & 15.0\% & 0.28 & 0.16 & 8.3\% & 0.18 & 0.10 & 0.0\% & 0.05 & 0.05 & 0.0\% & 0.06 & 0.06 \\
    \midrule
    \rowcolor{gray!15}Relational & 42 & 33.8\% & 0.47 & 0.20 & 23.3\% & 0.37 & 0.17 & 10.0\% & 0.23 & 0.15 & 6.4\% & 0.19 & 0.13 & 5.7\% & 0.16 & 0.11 \\
    \quad conjunction & 8 & 43.8\% & 0.55 & 0.21 & 27.5\% & 0.38 & 0.14 & 10.0\% & 0.14 & 0.05 & 12.5\% & 0.21 & 0.10 & 3.8\% & 0.06 & 0.03 \\
    \quad counting & 7 & 65.7\% & 0.75 & 0.28 & 44.3\% & 0.59 & 0.27 & 18.6\% & 0.34 & 0.18 & 11.4\% & 0.28 & 0.18 & 22.9\% & 0.36 & 0.16 \\
    \quad spatial & 29 & 21.0\% & 0.36 & 0.19 & 15.9\% & 0.30 & 0.17 & 7.2\% & 0.22 & 0.16 & 3.1\% & 0.16 & 0.13 & 1.7\% & 0.12 & 0.11 \\
    \midrule
    \rowcolor{gray!15}Visual & 84 & 23.8\% & 0.39 & 0.20 & 8.8\% & 0.20 & 0.12 & 5.0\% & 0.15 & 0.11 & 1.9\% & 0.08 & 0.06 & 2.7\% & 0.09 & 0.07 \\
    \quad color & 26 & 23.8\% & 0.42 & 0.23 & 6.2\% & 0.17 & 0.11 & 3.8\% & 0.16 & 0.13 & 1.2\% & 0.08 & 0.07 & 0.8\% & 0.09 & 0.09 \\
    \quad semantics & 60 & 23.2\% & 0.39 & 0.20 & 10.5\% & 0.22 & 0.13 & 6.8\% & 0.17 & 0.10 & 2.3\% & 0.09 & 0.07 & 3.5\% & 0.09 & 0.06 \\
    \quad size & 6 & 30.0\% & 0.36 & 0.09 & 10.0\% & 0.18 & 0.09 & 0.0\% & 0.03 & 0.03 & 1.7\% & 0.03 & 0.01 & 0.0\% & 0.05 & 0.05 \\
    \bottomrule
    \end{tabular}
    }

    \end{table*}

\begin{table*}
    \centering
    \caption{\small \textbf{Trajectory-quality metrics} per task category on \framework. ``SPARC'' is the spectral arc length of the end-effector velocity profile (Sec.~\ref{sec:metrics})---smoother motions yield values closer to zero, jerkier motions are more negative. ``Speed'' is the mean end-effector speed in cm/s. Standard deviations across episodes are shown in gray. Capability metrics for the same categories are reported in Table~\ref{table:categories_details}.}
    \label{table:categories_details_traj}
    \resizebox{\textwidth}{!}{%
    \begin{tabular}{lrrrrrrrrrrrr}
    \toprule
    \rowcolor{nvidiagreen!15} \multirow{2}{*}{\shortstack{\textbf{Task}\\\textbf{Categories}}} & \multirow{2}{*}{\textbf{\#}} & \multicolumn{2}{c}{\textbf{\piZeroFive}} & \multicolumn{2}{c}{\textbf{\piZeroFast}} & \multicolumn{2}{c}{\textbf{\groot}} & \multicolumn{2}{c}{\textbf{\piZero}} & \multicolumn{2}{c}{\textbf{\paliGemma}} \\
    \cmidrule(lr){3-4} \cmidrule(lr){5-6} \cmidrule(lr){7-8} \cmidrule(lr){9-10} \cmidrule(lr){11-12}
     &  & SPARC & Speed (cm/s) & SPARC & Speed (cm/s) & SPARC & Speed (cm/s) & SPARC & Speed (cm/s) & SPARC & Speed (cm/s) \\
    \midrule
    \rowcolor{gray!15}Total & 120 & $-$8.34 {\scriptsize\textcolor{gray}{($\pm$6.65)}} & 5.4 {\scriptsize\textcolor{gray}{($\pm$1.6)}} & $-$9.63 {\scriptsize\textcolor{gray}{($\pm$5.40)}} & 4.6 {\scriptsize\textcolor{gray}{($\pm$1.8)}} & $-$6.87 {\scriptsize\textcolor{gray}{($\pm$4.63)}} & 4.3 {\scriptsize\textcolor{gray}{($\pm$1.4)}} & $-$9.49 {\scriptsize\textcolor{gray}{($\pm$4.12)}} & 4.2 {\scriptsize\textcolor{gray}{($\pm$1.3)}} & $-$16.52 {\scriptsize\textcolor{gray}{($\pm$10.21)}} & 1.9 {\scriptsize\textcolor{gray}{($\pm$1.6)}} \\
    \quad simple & 64 & $-$7.38 {\scriptsize\textcolor{gray}{($\pm$6.66)}} & 5.4 {\scriptsize\textcolor{gray}{($\pm$1.4)}} & $-$8.09 {\scriptsize\textcolor{gray}{($\pm$4.41)}} & 4.7 {\scriptsize\textcolor{gray}{($\pm$1.7)}} & $-$6.37 {\scriptsize\textcolor{gray}{($\pm$5.00)}} & 4.5 {\scriptsize\textcolor{gray}{($\pm$1.4)}} & $-$8.46 {\scriptsize\textcolor{gray}{($\pm$4.37)}} & 4.3 {\scriptsize\textcolor{gray}{($\pm$1.4)}} & $-$15.33 {\scriptsize\textcolor{gray}{($\pm$8.73)}} & 2.0 {\scriptsize\textcolor{gray}{($\pm$1.7)}} \\
    \quad moderate & 39 & $-$8.85 {\scriptsize\textcolor{gray}{($\pm$7.21)}} & 5.4 {\scriptsize\textcolor{gray}{($\pm$1.8)}} & $-$10.07 {\scriptsize\textcolor{gray}{($\pm$5.44)}} & 4.7 {\scriptsize\textcolor{gray}{($\pm$2.1)}} & $-$6.92 {\scriptsize\textcolor{gray}{($\pm$4.21)}} & 4.2 {\scriptsize\textcolor{gray}{($\pm$1.3)}} & $-$10.13 {\scriptsize\textcolor{gray}{($\pm$2.99)}} & 4.1 {\scriptsize\textcolor{gray}{($\pm$1.3)}} & $-$15.02 {\scriptsize\textcolor{gray}{($\pm$7.52)}} & 2.1 {\scriptsize\textcolor{gray}{($\pm$1.7)}} \\
    \quad complex & 17 & $-$10.83 {\scriptsize\textcolor{gray}{($\pm$3.89)}} & 4.8 {\scriptsize\textcolor{gray}{($\pm$1.3)}} & $-$14.38 {\scriptsize\textcolor{gray}{($\pm$5.75)}} & 4.1 {\scriptsize\textcolor{gray}{($\pm$1.3)}} & $-$8.63 {\scriptsize\textcolor{gray}{($\pm$3.52)}} & 4.1 {\scriptsize\textcolor{gray}{($\pm$1.4)}} & $-$11.91 {\scriptsize\textcolor{gray}{($\pm$4.09)}} & 3.8 {\scriptsize\textcolor{gray}{($\pm$1.1)}} & $-$24.46 {\scriptsize\textcolor{gray}{($\pm$15.68)}} & 1.5 {\scriptsize\textcolor{gray}{($\pm$0.8)}} \\
    \midrule
    \rowcolor{gray!15}Procedural & 22 & $-$10.08 {\scriptsize\textcolor{gray}{($\pm$4.93)}} & 4.8 {\scriptsize\textcolor{gray}{($\pm$1.4)}} & $-$11.06 {\scriptsize\textcolor{gray}{($\pm$6.31)}} & 4.3 {\scriptsize\textcolor{gray}{($\pm$1.4)}} & $-$6.98 {\scriptsize\textcolor{gray}{($\pm$5.16)}} & 3.8 {\scriptsize\textcolor{gray}{($\pm$1.3)}} & $-$10.58 {\scriptsize\textcolor{gray}{($\pm$2.96)}} & 4.1 {\scriptsize\textcolor{gray}{($\pm$1.1)}} & $-$16.07 {\scriptsize\textcolor{gray}{($\pm$7.69)}} & 1.5 {\scriptsize\textcolor{gray}{($\pm$0.8)}} \\
    \quad affordance & 12 & $-$10.80 {\scriptsize\textcolor{gray}{($\pm$4.47)}} & 4.1 {\scriptsize\textcolor{gray}{($\pm$1.3)}} & $-$11.74 {\scriptsize\textcolor{gray}{($\pm$5.97)}} & 3.6 {\scriptsize\textcolor{gray}{($\pm$1.0)}} & $-$6.79 {\scriptsize\textcolor{gray}{($\pm$2.49)}} & 3.9 {\scriptsize\textcolor{gray}{($\pm$1.2)}} & $-$11.12 {\scriptsize\textcolor{gray}{($\pm$3.20)}} & 4.0 {\scriptsize\textcolor{gray}{($\pm$1.2)}} & $-$15.40 {\scriptsize\textcolor{gray}{($\pm$8.50)}} & 1.5 {\scriptsize\textcolor{gray}{($\pm$0.8)}} \\
    \quad reorientation & 6 & $-$10.08 {\scriptsize\textcolor{gray}{($\pm$3.69)}} & 4.8 {\scriptsize\textcolor{gray}{($\pm$1.2)}} & $-$11.87 {\scriptsize\textcolor{gray}{($\pm$7.80)}} & 4.4 {\scriptsize\textcolor{gray}{($\pm$1.4)}} & $-$7.05 {\scriptsize\textcolor{gray}{($\pm$1.64)}} & 3.2 {\scriptsize\textcolor{gray}{($\pm$0.8)}} & $-$10.97 {\scriptsize\textcolor{gray}{($\pm$2.64)}} & 3.9 {\scriptsize\textcolor{gray}{($\pm$0.9)}} & $-$17.63 {\scriptsize\textcolor{gray}{($\pm$8.49)}} & 1.3 {\scriptsize\textcolor{gray}{($\pm$0.5)}} \\
    \quad stacking & 6 & $-$9.04 {\scriptsize\textcolor{gray}{($\pm$5.99)}} & 5.9 {\scriptsize\textcolor{gray}{($\pm$1.1)}} & $-$9.05 {\scriptsize\textcolor{gray}{($\pm$4.12)}} & 5.4 {\scriptsize\textcolor{gray}{($\pm$1.2)}} & $-$7.41 {\scriptsize\textcolor{gray}{($\pm$9.13)}} & 4.3 {\scriptsize\textcolor{gray}{($\pm$1.5)}} & $-$9.35 {\scriptsize\textcolor{gray}{($\pm$2.28)}} & 4.1 {\scriptsize\textcolor{gray}{($\pm$0.9)}} & $-$16.72 {\scriptsize\textcolor{gray}{($\pm$6.59)}} & 1.4 {\scriptsize\textcolor{gray}{($\pm$0.9)}} \\
    \midrule
    \rowcolor{gray!15}Relational & 42 & $-$7.50 {\scriptsize\textcolor{gray}{($\pm$3.95)}} & 5.5 {\scriptsize\textcolor{gray}{($\pm$1.6)}} & $-$8.70 {\scriptsize\textcolor{gray}{($\pm$4.81)}} & 4.9 {\scriptsize\textcolor{gray}{($\pm$2.1)}} & $-$6.51 {\scriptsize\textcolor{gray}{($\pm$2.48)}} & 4.3 {\scriptsize\textcolor{gray}{($\pm$1.3)}} & $-$8.87 {\scriptsize\textcolor{gray}{($\pm$2.92)}} & 4.5 {\scriptsize\textcolor{gray}{($\pm$1.4)}} & $-$14.93 {\scriptsize\textcolor{gray}{($\pm$10.38)}} & 2.4 {\scriptsize\textcolor{gray}{($\pm$2.1)}} \\
    \quad conjunction & 8 & $-$6.88 {\scriptsize\textcolor{gray}{($\pm$4.04)}} & 5.7 {\scriptsize\textcolor{gray}{($\pm$1.4)}} & $-$7.96 {\scriptsize\textcolor{gray}{($\pm$3.97)}} & 5.0 {\scriptsize\textcolor{gray}{($\pm$1.6)}} & $-$7.47 {\scriptsize\textcolor{gray}{($\pm$3.05)}} & 3.6 {\scriptsize\textcolor{gray}{($\pm$1.2)}} & $-$7.74 {\scriptsize\textcolor{gray}{($\pm$2.07)}} & 4.8 {\scriptsize\textcolor{gray}{($\pm$1.2)}} & $-$13.77 {\scriptsize\textcolor{gray}{($\pm$9.36)}} & 2.4 {\scriptsize\textcolor{gray}{($\pm$1.9)}} \\
    \quad counting & 7 & $-$7.16 {\scriptsize\textcolor{gray}{($\pm$4.91)}} & 6.7 {\scriptsize\textcolor{gray}{($\pm$2.3)}} & $-$9.43 {\scriptsize\textcolor{gray}{($\pm$5.95)}} & 5.9 {\scriptsize\textcolor{gray}{($\pm$3.1)}} & $-$7.64 {\scriptsize\textcolor{gray}{($\pm$2.62)}} & 4.0 {\scriptsize\textcolor{gray}{($\pm$1.2)}} & $-$9.61 {\scriptsize\textcolor{gray}{($\pm$2.60)}} & 4.2 {\scriptsize\textcolor{gray}{($\pm$1.0)}} & $-$9.88 {\scriptsize\textcolor{gray}{($\pm$5.49)}} & 3.7 {\scriptsize\textcolor{gray}{($\pm$2.7)}} \\
    \quad spatial & 29 & $-$8.17 {\scriptsize\textcolor{gray}{($\pm$3.90)}} & 5.0 {\scriptsize\textcolor{gray}{($\pm$1.3)}} & $-$8.99 {\scriptsize\textcolor{gray}{($\pm$4.76)}} & 4.5 {\scriptsize\textcolor{gray}{($\pm$1.8)}} & $-$6.06 {\scriptsize\textcolor{gray}{($\pm$2.06)}} & 4.6 {\scriptsize\textcolor{gray}{($\pm$1.3)}} & $-$9.18 {\scriptsize\textcolor{gray}{($\pm$3.16)}} & 4.5 {\scriptsize\textcolor{gray}{($\pm$1.5)}} & $-$16.42 {\scriptsize\textcolor{gray}{($\pm$10.84)}} & 2.0 {\scriptsize\textcolor{gray}{($\pm$1.8)}} \\
    \midrule
    \rowcolor{gray!15}Visual & 84 & $-$8.73 {\scriptsize\textcolor{gray}{($\pm$7.39)}} & 5.3 {\scriptsize\textcolor{gray}{($\pm$1.6)}} & $-$10.52 {\scriptsize\textcolor{gray}{($\pm$5.60)}} & 4.3 {\scriptsize\textcolor{gray}{($\pm$1.7)}} & $-$7.10 {\scriptsize\textcolor{gray}{($\pm$4.71)}} & 4.3 {\scriptsize\textcolor{gray}{($\pm$1.3)}} & $-$9.81 {\scriptsize\textcolor{gray}{($\pm$3.49)}} & 3.9 {\scriptsize\textcolor{gray}{($\pm$1.2)}} & $-$17.72 {\scriptsize\textcolor{gray}{($\pm$10.98)}} & 1.8 {\scriptsize\textcolor{gray}{($\pm$1.4)}} \\
    \quad color & 26 & $-$7.93 {\scriptsize\textcolor{gray}{($\pm$4.31)}} & 5.2 {\scriptsize\textcolor{gray}{($\pm$1.5)}} & $-$9.75 {\scriptsize\textcolor{gray}{($\pm$4.91)}} & 4.3 {\scriptsize\textcolor{gray}{($\pm$1.3)}} & $-$6.74 {\scriptsize\textcolor{gray}{($\pm$5.72)}} & 4.1 {\scriptsize\textcolor{gray}{($\pm$1.3)}} & $-$9.30 {\scriptsize\textcolor{gray}{($\pm$2.83)}} & 3.9 {\scriptsize\textcolor{gray}{($\pm$1.2)}} & $-$16.43 {\scriptsize\textcolor{gray}{($\pm$10.28)}} & 2.0 {\scriptsize\textcolor{gray}{($\pm$1.5)}} \\
    \quad semantics & 60 & $-$9.26 {\scriptsize\textcolor{gray}{($\pm$8.44)}} & 5.4 {\scriptsize\textcolor{gray}{($\pm$1.6)}} & $-$10.96 {\scriptsize\textcolor{gray}{($\pm$5.74)}} & 4.2 {\scriptsize\textcolor{gray}{($\pm$1.8)}} & $-$7.32 {\scriptsize\textcolor{gray}{($\pm$5.35)}} & 4.4 {\scriptsize\textcolor{gray}{($\pm$1.3)}} & $-$10.18 {\scriptsize\textcolor{gray}{($\pm$3.70)}} & 4.0 {\scriptsize\textcolor{gray}{($\pm$1.3)}} & $-$17.99 {\scriptsize\textcolor{gray}{($\pm$11.20)}} & 1.8 {\scriptsize\textcolor{gray}{($\pm$1.4)}} \\
    \quad size & 6 & $-$6.85 {\scriptsize\textcolor{gray}{($\pm$2.34)}} & 5.4 {\scriptsize\textcolor{gray}{($\pm$1.4)}} & $-$7.82 {\scriptsize\textcolor{gray}{($\pm$3.53)}} & 4.4 {\scriptsize\textcolor{gray}{($\pm$1.3)}} & $-$7.30 {\scriptsize\textcolor{gray}{($\pm$2.20)}} & 4.3 {\scriptsize\textcolor{gray}{($\pm$1.7)}} & $-$7.77 {\scriptsize\textcolor{gray}{($\pm$2.76)}} & 3.7 {\scriptsize\textcolor{gray}{($\pm$1.1)}} & $-$17.76 {\scriptsize\textcolor{gray}{($\pm$7.61)}} & 1.7 {\scriptsize\textcolor{gray}{($\pm$0.9)}} \\
    \bottomrule
    \end{tabular}
    }

    \end{table*}

\section{Details of MNPE Sensitivity Analysis} \label{app:mnpe}

MNPE allows us to analyze the relationship between scene parameters and policy outcomes in a likelihood-free Bayesian inference setting.

\paragraph{Variable Definitions}
Let $\theta \in \Theta$ denote the vector of variation parameters. In our camera pose sensitivity experiments, 
$
\theta = (d_{\text{ext}}, d_{\text{wrist}}) \in \mathbb{R}^2
$
where $d_{\text{ext}}$ and $d_{\text{wrist}}$ represent the displacement of the external and wrist cameras from their reference configurations, respectively, in SE(3).

Let $x \in \{0, 1\}$ denote the binary task success indicator, and let $\pi$ denote the robot policy being evaluated.

\paragraph{Handling Mixed Parameters}
For experiments involving both continuous parameters (e.g., pose distances) and discrete parameters (e.g., lighting levels, table materials), MNPE handles mixed continuous-discrete parameters through factorization: $q_\phi(\theta \mid x) = q_\phi(\theta^{\text{cont}} \mid \theta^\text{disc}, x) \cdot q_\phi(\theta^\text{disc} \mid x)$, where discrete components use softmax distributions and continuous components use normalizing flows. 
In our camera pose experiments, all parameters are continuous.

\paragraph{Pose Distance Metric}
Poses are represented as 7-DoF transformations $\mathbf{T} = (\mathbf{p}, \mathbf{q})$ comprising position $\mathbf{p} \in \mathbb{R}^3$ and unit quaternion orientation $\mathbf{q} \in \mathbb{H}$. We compute a weighted distance from the reference configuration:
\begin{equation}
d(\mathbf{T}, \mathbf{T}_{\text{ref}}) = \|\mathbf{p} - \mathbf{p}_{\text{ref}}\|_2 + \beta \cdot d_{\text{SO(3)}}(\mathbf{q}, \mathbf{q}_{\text{ref}}),
\end{equation}
where the geodesic distance on SO(3) is:
\begin{equation}
d_{\text{SO(3)}}(\mathbf{q}_1, \mathbf{q}_2) = 2 \arccos\left(\min(1, |\mathbf{q}_1 \cdot \mathbf{q}_2|)\right).
\end{equation}
The weighting factor $\beta = 1.0$ balances translational (meters) and rotational (radians) components. 
For camera displacement, reference poses correspond to nominal camera mounting positions.
For object pose, reference pose is the origin of the robot base.

\paragraph{Prior Specification}
We adopt non-informative uniform priors to avoid biasing the inference toward any particular parameter region. For continuous parameters normalized to the unit interval:
\begin{equation}
p(\theta) = \prod_{j=1}^{m} \text{Uniform}(0, 1) = 1, \quad \theta_j \in [0, 1].
\end{equation}

\paragraph{Training Objective}
The neural network parameters $\phi$ are optimized by minimizing the negative log-likelihood over the training dataset, $\mathcal{D} = \{(\theta_i, x_i)\}_{i=1}^{N}$:
\begin{equation}
\mathcal{L}(\phi) = -\frac{1}{N} \sum_{i=1}^{N} \log q_\phi(\theta_i \mid x_i).
\end{equation}
We train for 50 epochs using the Adam optimizer on data collected from the camera pose variation and initial pose variation experiments (see Table~\ref{table:variations}).

\paragraph{Importance Sampling Correction}
Since the experimental data may sample parameters non-uniformly, we apply importance sampling to recover the posterior under a uniform prior: $p(\theta \mid x) \approx \frac{p(\theta)}{\tilde{p}(\theta)} q_\phi(\theta \mid x)$, where $\tilde{p}(\theta)$ is the empirical proposal distribution. 
We correct posterior samples using importance weights:
\begin{equation}
w_i = \frac{p(\theta_i)}{\tilde{p}(\theta_i)},
\end{equation}
where $\tilde{p}(\theta)$ is estimated via Gaussian kernel density estimation on the training data. The effective sample size $\text{ESS} = 1 / \sum_i \bar{w}_i^2$ quantifies the efficiency of this correction.

\paragraph{Posterior Inference}
Given a query observation $x_o$ (e.g., $x_o = 1$ for successful task completion), we draw $N_s = 5000$ samples from the learned posterior:
\begin{equation}
\left\{ \theta^{(i)} \right\}_{i=1}^{N_s} \sim q_\phi(\theta \mid x_o).
\end{equation}

\paragraph{Posterior Statistics}
For each continuous parameter, we compute the posterior mean and 95\% credible interval:
\begin{align}
\hat{\mu}_j &= \frac{1}{N_s} \sum_{i=1}^{N_s} \theta_j^{(i)}, \\
\text{CI}_{95\%}^{(j)} &= \left[ Q_{0.025}\left(\{\theta_j^{(i)}\}\right), \; Q_{0.975}\left(\{\theta_j^{(i)}\}\right) \right],
\end{align}
where $Q_\alpha(\cdot)$ denotes the $\alpha$-quantile.

This analysis reveals which variation parameters are most strongly associated with successful task outcomes: a posterior distribution tightly concentrated near zero indicates high sensitivity to that parameter (the policy requires it to remain near the reference value), while a broad posterior indicates robustness to variation.

\section{Details on Scaling Scene Generation}
\label{sec:scaling-scene-generation}

We present additional implementation details on scene generation (\Cref{sec:scenegen}).

\begin{figure}[bh]
    \centering
    \includegraphics{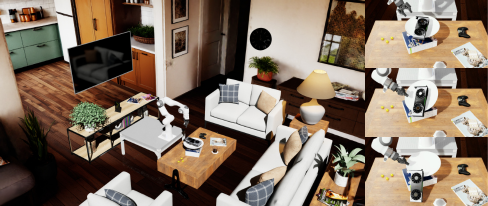}
    \caption{(\textit{Left}) We show one of our Gaussian Splat + Mesh scenes in \framework. This scene has a Gaussian splat background with a collision mesh for the splat estimated with 3DGRUT~\cite{loccoz20243dgrt, wu20253dgut}, and a mesh foreground. All objects in the scene have spatially varying density, and thus mass is estimated with VoMP~\cite{dagli2025vomp}. (\textit{Right}) We show a VLA running a task in this scene.}
    \label{fig:splats2}
\end{figure}

\subsection{Stage I: Predicates for Semantic Planning}
The following predicates are used:
\begin{itemize}
    \item $\mathtt{PlaceIn}(x, y)$: Object $x$ must be contained within $y$ (e.g., fruit in a bowl).
    \item $\mathtt{PlaceOn}(x, y)$: Object $x$ is supported by $y$ (e.g., mug on a coaster).
    \item $\mathtt{ClusterAround}(x, \{y_i\})$: Object $x$ acts as an anchor for a group $\{B_i\}$.
\item $\mathtt{PlaceAnywhere}(x)$: Object $x$ is placed freely on the global support surface (table).
  \end{itemize}
If a predicate refers to a non-existent anchor, it is downgraded to a $\mathtt{PlaceAnywhere}$ constraint to preserve the object in the scene.

\subsection{Stage II: Geometric Constraint Solving}

For \textbf{Global Placement} ($\mathtt{PlaceAnywhere}$), we utilize rejection sampling on the global table surface bounds, checking collision against all currently placed objects using SAT on OBBs. To handle high-density scenes, we employ two strategies: (1) an adaptive relaxation loop that progressively increases collision margins if a valid layout is not found, and (2) a stochastic perturbation step that randomly jitters all object positions when the solver converges to a local minimum (\Cref{alg:spatial}).

For \textbf{Stacking} ($\mathtt{PlaceOn}(b_i, b_{support})$), we sample positions on the top surface of $b_{support}$ using rejection sampling (up to $K=20$ attempts) to find a position $\mathbf{p}_{xy}$ such that $\text{OBB}(b_i) \cap \text{OBB}(b_{existing}) = \emptyset$ for all previously placed objects on the same support (\Cref{alg:physical}).

For \textbf{Containment} ($\mathtt{PlaceIn}(b_i, b_{container})$), we compute the available interior volume of $b_{container}$ using its bounding box dimensions $\mathbf{d}_b$. We employ a packing heuristic that discretizes the container's floor into a grid with resolution $s = \max(\mathbf{d}_i^{x}, \mathbf{d}_i^{y}) + \epsilon_{margin}$. A cell $(u, v)$ is considered valid if it is unoccupied and within the container's bounds scaled by factor $\gamma=0.7$ to avoid edge collisions. We assign $o_i$ to the first valid cell, setting its height $z_i = z_{container} + h_{container}/2$.

\begin{algorithm}[tb]
    \caption{Spatial Constraint Solver}
    \label{alg:spatial}
    \textbf{Input:} Objects $B$, Predicates $P$, Table Bounds $L_\text{max}$ \\
    \textbf{Output:} 2D coordinates $(x, y, \theta)$ for all base objects
    \begin{algorithmic}[1]
    \STATE \textbf{Margins} $M \gets [\mu, 1.25\mu, 1.5\mu, 2.0\mu]$
    \FORALL{$\text{margin} \in M$}
        \STATE \COMMENT{\textcolor{blue}{Phase 1: Initialization}}
        \STATE Randomize $(x,y)$ for all loose objects inside $L_\text{max}$
        \FORALL{$p \in P$}
            \IF{$p.\text{type} == \text{place-on-base}$}
                \STATE $p.\text{object}.(x,y,\theta) \gets (p.x, p.y, p.\text{yaw})$
            \ELSIF{$p.\text{type} == \text{cluster-around}$}
                \STATE \textbf{PolarPlace}($p.\text{targets}$, $p.\text{anchor}$, $p.\text{radius}$)
            \ENDIF
        \ENDFOR
        \STATE \COMMENT{\textcolor{blue}{Phase 2: Relative Constraints}}
        \WHILE{constraints not satisfied}
            \STATE \text{ApplyRelativeConstraints}($P$)
        \ENDWHILE
        \STATE \text{ApplyOrientations}($P$)
        \STATE \COMMENT{\textcolor{blue}{Phase 3: Collision Resolution}}
        \FOR{$k = 1$ \textbf{to} $K_{\text{max}}$}
            \STATE $C \gets \text{FindCollisions}(B, \text{margin})$
            \IF{$C = \emptyset$} \RETURN \textbf{Success} \ENDIF
            \IF{$|C|$ not decreasing for 10 steps}
                \STATE \textbf{PerturbPositions}(B)
            \ENDIF
            \FORALL{$(o_i, o_j) \in C$}
                \STATE \textbf{ResolveOverlap}($b_i, b_j, \text{margin}$)
                \STATE \textbf{ClampToBounds}($b_i, b_j, L_\text{max}$)
            \ENDFOR
        \ENDFOR
    \ENDFOR
    \RETURN \textbf{Failure}
    \end{algorithmic}
\end{algorithm}

\begin{figure*}[tb]
\centering
\begin{tcolorbox}[
    colback=blue!5!white,       %
    colframe=blue!50!black,     %
    arc=3mm,                    %
    boxrule=0.5pt,              %
    width=\textwidth,           %
    left=3mm, right=3mm,
    top=2mm, bottom=2mm
]
You are a scene generation expert creating REALISTIC robot manipulation scenarios.\\[0.5em]
\textbf{REAL-WORLD SCENE PRINCIPLES}:\\
1. Objects form CLUSTERS - not evenly spaced grids\\
2. Containers (bowls, bins) have objects INSIDE them\\
3. Supports (plates, trays) have objects ON TOP\\
4. Objects scatter naturally AROUND containers\\
5. Orientations VARY - not all aligned to 0$^\circ$/90$^\circ$\\[0.5em]
\textbf{COORDINATE SYSTEM}:\\
- Table bounds: X=[0.25 to 0.85], Y=[-0.40 to 0.40] (meters)\\
- Table center: (0.55, 0.0)\\
- Front=+X, Back=-X, Left=+Y, Right=-Y\\[0.5em]
\textbf{PLACEMENT TYPES}:\\
1. \textbf{place-on-base}: Object directly on table\\
\quad\{``type'': ``place-on-base'', ``object'': ``bowl\_0'', ``x'': 0.4, ``y'': 0.1, ``yaw'': 23\}\\
\quad VARY yaw angles (15, 47, 123, not just 0/90/180).\\
\quad Position matters for anchors, less for loose objects.\\
2. \textbf{place-in}: Objects INSIDE a container\\
\quad\{``type'': ``place-in'', ``objects'': [``apple\_01'', ``orange\_01''], ``container'': ``bowl\_0''\}\\
\quad Container MUST be placed first with place-on-base.\\
\quad Great for fruits in bowls, items in bins.\\
3. \textbf{place-on}: Object ON TOP of support\\
\quad\{``type'': ``place-on'', ``object'': ``banana'', ``support'': ``plate\_large'', ``position'': ``center''\}\\
\quad Support MUST be placed first.\\
\quad position: ``center'', ``edge'', or ``random''\\
\quad Great for food on plates, items on trays.\\
4. \textbf{cluster-around}: Objects scattered NEAR an anchor\\
\quad\{``type'': ``cluster-around'', ``objects'': [``mug'', ``spoon''], ``anchor'': ``bowl\_0'', ``radius'': 0.15\}\\
\quad Creates natural groupings.\\[0.5em]
\quad radius: how far from anchor (0.10--0.20m typical)\\[0.5em]
\textbf{SCENE STRUCTURE} (follow this pattern):\\
1. Place 1-2 ANCHOR objects (containers/supports) on table\\
2. Put objects INSIDE containers (place-in)\\
3. Put objects ON supports (place-on)\\
4. Cluster objects AROUND anchors (cluster-around)\\
5. Add a few LOOSE objects to fill space\\[0.5em]
\textbf{REALISTIC SPACING}:\\
- Anchors: 0.25-0.35m apart\\
- Clustered objects: 0.08-0.15m from anchor\\
- Loose objects: fill remaining space naturally
\end{tcolorbox}
\caption{System prompt for Stage I (Semantic Planning). This prompt instructs the LLM to generate physically plausible scene layouts using structured predicates rather than raw coordinates.}
\label{fig:system_prompt1}
\end{figure*}

\begin{figure*}[tb]
\centering
\begin{tcolorbox}[
    colback=blue!5!white,       %
    colframe=blue!50!black,     %
    arc=3mm,                    %
    boxrule=0.5pt,              %
    width=\textwidth,           %
    left=3mm, right=3mm,
    top=2mm, bottom=2mm
]
\textbf{OUTPUT FORMAT} (JSON only, no markdown):\\
\{\\
\quad ``objects'': [\\
\quad\quad \{``name'': ``bowl\_0''\},\\
\quad\quad \{``name'': ``plate\_large''\},\\
\quad\quad \{``name'': ``apple\_01''\},\\
\quad\quad \{``name'': ``orange\_01''\},\\
\quad\quad \{``name'': ``banana''\},\\
\quad\quad \{``name'': ``mug''\},\\
\quad\quad \{``name'': ``spoon''\}\\
\quad ],\\
\quad ``predicates'': [\\
\quad\quad \{``type'': ``place-on-base'', ``object'': ``bowl\_0'', ``x'': 0.40, ``y'': 0.15, ``yaw'': 23\},\\
\quad\quad \{``type'': ``place-on-base'', ``object'': ``plate\_large'', ``x'': 0.65, ``y'': -0.10, ``yaw'': 156\},\\
\quad\quad \{``type'': ``place-in'', ``objects'': [``apple\_01'', ``orange\_01''], ``container'': ``bowl\_0''\},\\
\quad\quad \{``type'': ``place-on'', ``object'': ``banana'', ``support'': ``plate\_large'', ``position'': ``center''\},\\
\quad\quad \{``type'': ``cluster-around'', ``objects'': [``mug'', ``spoon''], ``anchor'': ``bowl\_0'', ``radius'': 0.12\}\\
\quad ]\\
\}\\[0.5em]
\textbf{CRITICAL RULES}:\\
1. Object names MUST match EXACTLY from catalog\\
2. Containers/supports MUST be placed before objects go in/on them\\
3. Create INTERESTING scenes with containment, stacking, AND clustering\\
4. VARY yaw angles - real scenes aren't grid-aligned\\
5. Return ONLY valid JSON, no markdown
\end{tcolorbox}
\caption{Continued System prompt for Stage I (Semantic Planning). This prompt instructs the LLM to generate physically plausible scene layouts using structured predicates rather than raw coordinates.}
\label{fig:system_prompt2}
\end{figure*}

\begin{figure*}[tb]
\centering
\begin{tcolorbox}[
    colback=blue!5!white,       %
    colframe=blue!50!black,     %
    arc=3mm,                    %
    boxrule=0.5pt,              %
    width=\textwidth,           %
    left=3mm, right=3mm,
    top=2mm, bottom=2mm
]
\textbf{SCENE REQUEST}: \colorbox{yellow!30}{theme from dataset}\\
\textbf{TARGET}: \colorbox{yellow!30}{target object count} objects\\[0.5em]
\textbf{TABLE SIZE}: 0.7m $\times$ 1.0m = 0.70m$^2$ (objects must fit with spacing!)\\[0.5em]
\textbf{SIZE LIMITS} (max 1-2 large objects, prefer smaller for 8+ items):\\
\quad Large (\(>0.08\mathrm{m}^2\)): \colorbox{yellow!30}{computed from catalog footprint}\\
\quad Avoid picking multiple large objects - they won't all fit!\\[0.5em]
\textbf{AVAILABLE OBJECTS}:\\
\textbf{CONTAINERS} (can hold objects inside): \colorbox{yellow!30}{filled from catalog}\\
\textbf{SUPPORTS} (can stack objects on): \colorbox{yellow!30}{filled from catalog}\\
\textbf{FOOD}: \colorbox{yellow!30}{filled from catalog}\\
\textbf{TOOLS}: \colorbox{yellow!30}{filled from catalog}\\
\textbf{OTHER}: \colorbox{yellow!30}{filled from catalog}\\[0.5em]
\textbf{MEDIUM SCENE STRATEGY} (10-14 objects):\\
- Use 1-2 containers/supports as ANCHORS\\
- Put 2-4 objects IN containers (place-in)\\
- Stack 1-2 items ON supports (place-on)\\
- Cluster 2-3 objects near anchors (cluster-around)\\
- VARY yaw angles - no grid alignment!\\[0.5em]
\textbf{SUGGESTED for diversity} (use only if they fit the theme): \colorbox{yellow!30}{preselected objects}
\end{tcolorbox}
\caption{User prompt template for Stage I (medium target count). The highlighted fields are populated at runtime (theme, target count, size warnings, catalog subsets, and diversity suggestions). Analogous strategy blocks are used for sparse (fewer than 10) and dense (15+) targets.}
\label{fig:user_prompt}
\end{figure*}

\begin{figure*}[tb]
\centering
\begin{tcolorbox}[
    colback=red!5!white,
    colframe=red!50!black,
    arc=3mm,
    boxrule=0.5pt,
    width=\textwidth,
    left=3mm, right=3mm,
    top=2mm, bottom=2mm
]
\textbf{PREVIOUS ATTEMPT FAILED}:\\
\colorbox{yellow!30}{feedback string produced by spatial/physical solver or grammar checks}\\[0.5em]
Please fix the issues. Common fixes:\\
- Use MORE containment (place-in) to reduce table crowding\\
- Use MORE stacking (place-on) to utilize vertical space\\
- Use clustering (cluster-around) instead of individual placements\\
- Select SMALLER objects if collisions persist
\end{tcolorbox}
\caption{Feedback block appended to the user prompt when spatial solving, physical placement, grammar checks, or intersection validation fails. The highlighted region is the dynamic diagnostic message.}
\label{fig:feedback_prompt}
\end{figure*}

\subsection{Baseline Method}
\label{sec:baseline}

To validate the efficacy of our hierarchical approach, we implement a robust baseline inspired by standard domain randomization techniques. The baseline operates in a single pass without iterative feedback. The process begins with the LLM selecting a list of objects $O$ and suggesting a grid layout (rows $R \times$ columns $C$) for the table surface. The table surface is then divided into $R \times C$ rectangular cells, and objects are assigned to cells sequentially. Within each cell $k$, the object's position is jittered uniformly: $\mathbf{p}_{xy} \sim \mathcal{U}(\text{center}_k - w/4, \text{center}_k + w/4)$.
This ensures basic separation but precludes complex stacking or containment, as objects are simply placed at a safe height $z = z_{table} + h_{obj}/2$. Finally, we run the same physics simulation pass as in our method to allow objects to settle under gravity, resolving minor inter-penetrations but without the capability to correct semantic or structural failures.

\subsection{Experiments}

We compare our scene generation with the baseline method using popular scene generation metrics, VQA score~\cite{lin2024evaluatingtexttovisualgenerationimagetotext}, GPT preference where it is shown two images each from the baseline or our method and is asked to pick one, and following~\citet{yang2025sceneweaverallinone3dscene}, we report the visual realism (Real.), functionality (Func.), layout correctness (Lay.), Quality (Qual.) and scene completeness (Comp.) scores. We use GPT-4o~\cite{openai2024gpt4technicalreport} to generate scenes from our method and baseline; and use GPT-4.1~\cite{openai2024gpt4technicalreport} for the evaluations. These metrics are computed on rendered RGB images from two viewpoints: a frontal view aligned with the table axis (camera at $(1.0, 0.0, 0.7)$ looking at table center) and an angled perspective view (camera at $(-0.3, 0.3, 0.7)$).

We show quantitative comparisons across 100 generated scenes for our method compared to the baselines in~\Cref{tab:scene_gen:results_overall}. We show the quantitative comparisons split by the number of objects in the scene ($[1, 5]$ objects, $[6, 15]$ objects, and $[16, 20]$ objects) in~\Cref{tab:scene_gen:results_difficulty}. We show the quantitative comparisons split across the 10 scene themes we use in~\Cref{tab:scene_gen:results_per_scene}. We find our method consistently outperforms the baseline across all metrics, with particularly large gains in visual realism and semantic functionality.

\begin{table*}[tb]
    \centering
    \caption{\textbf{Quantitative comparison for Scene Generation.} We evaluate our against the baseline across diverse metrics measuring the visual realism (Real.), functionality (Func.), layout correctness (Lay.), Quality (Qual.), VQA score~\cite{lin2024evaluatingtexttovisualgenerationimagetotext}, and GPT Preference.}
    \label{tab:scene_gen:results_overall}
    \resizebox{\textwidth}{!}{%
    \begin{tabular}{lrrrrrrrr}
    \toprule
    \rowcolor{nvidiagreen!15}Method & VQA ($\uparrow$) & Real. ($\uparrow$) & Func. ($\uparrow$) & Lay. ($\uparrow$) & Compl. ($\uparrow$) & Qual. ($\uparrow$) & \# Obj ($\uparrow$) & GPT Pref. ($\uparrow$)\\
    \midrule
    Baseline & \underline{0.398} {\textbf{\scriptsize\textcolor{gray}{($\pm$0.04)}}} & \underline{6.889} {\textbf{\scriptsize\textcolor{gray}{($\pm$0.36)}}} & \underline{6.221} {\textbf{\scriptsize\textcolor{gray}{($\pm$0.58)}}} & \underline{6.166} {\textbf{\scriptsize\textcolor{gray}{($\pm$0.42)}}} & \underline{4.687} {\textbf{\scriptsize\textcolor{gray}{($\pm$0.57)}}} & \underline{5.991} {\textbf{\scriptsize\textcolor{gray}{($\pm$0.24)}}} & \underline{13.750} {\textbf{\scriptsize\textcolor{gray}{($\pm$10.52)}}} & \underline{18.000}\\
    Ours & \textbf{0.554} {\textbf{\scriptsize\textcolor{gray}{($\pm$0.03)}}} & \textbf{8.755} {\textbf{\scriptsize\textcolor{gray}{($\pm$0.27)}}} & \textbf{8.951} {\textbf{\scriptsize\textcolor{gray}{($\pm$0.33)}}} & \textbf{7.919} {\textbf{\scriptsize\textcolor{gray}{($\pm$0.28)}}} & \textbf{8.207} {\textbf{\scriptsize\textcolor{gray}{($\pm$0.40)}}} & \textbf{8.458} {\textbf{\scriptsize\textcolor{gray}{($\pm$0.25)}}} & \textbf{26.870} {\textbf{\scriptsize\textcolor{gray}{($\pm$24.90)}}} & \textbf{82.000}\\
    \bottomrule
    \end{tabular}%
    }
    \end{table*}
    
    \begin{table*}[tb]
    \centering
    \caption{\textbf{Quantitative comparison across Difficulty Splits.} We evaluate our method against the baseline on Easy, Medium, and Hard splits. Our method consistently outperforms the baseline across all difficulty levels.}
    \label{tab:scene_gen:results_difficulty}
    \resizebox{\textwidth}{!}{%
    \begin{tabular}{lrrrrrrrr}
    \toprule
    \rowcolor{nvidiagreen!15}Method & VQA ($\uparrow$) & Real. ($\uparrow$) & Func. ($\uparrow$) & Lay. ($\uparrow$) & Compl. ($\uparrow$) & Qual. ($\uparrow$) & \# Obj ($\uparrow$) & GPT Pref. ($\uparrow$)\\
    
    \midrule
    \rowcolor{gray!15}\multicolumn{9}{l}{Easy ($[0,5]$ objects)}\\
    \midrule
    Baseline & \underline{0.458} {\textbf{\scriptsize\textcolor{gray}{($\pm$0.02)}}} & \underline{6.767} {\textbf{\scriptsize\textcolor{gray}{($\pm$0.36)}}} & \underline{6.269} {\textbf{\scriptsize\textcolor{gray}{($\pm$0.57)}}} & \underline{6.079} {\textbf{\scriptsize\textcolor{gray}{($\pm$0.48)}}} & \underline{4.737} {\textbf{\scriptsize\textcolor{gray}{($\pm$0.61)}}} & \underline{5.963} {\textbf{\scriptsize\textcolor{gray}{($\pm$0.26)}}} & \underline{6.467} {\textbf{\scriptsize\textcolor{gray}{($\pm$0.52)}}} & \underline{33.333}\\
    Ours & \textbf{0.525} {\textbf{\scriptsize\textcolor{gray}{($\pm$0.05)}}} & \textbf{8.331} {\textbf{\scriptsize\textcolor{gray}{($\pm$0.24)}}} & \textbf{8.423} {\textbf{\scriptsize\textcolor{gray}{($\pm$0.27)}}} & \textbf{7.636} {\textbf{\scriptsize\textcolor{gray}{($\pm$0.21)}}} & \textbf{7.626} {\textbf{\scriptsize\textcolor{gray}{($\pm$0.25)}}} & \textbf{8.004} {\textbf{\scriptsize\textcolor{gray}{($\pm$0.09)}}} & \textbf{12.800} {\textbf{\scriptsize\textcolor{gray}{($\pm$11.54)}}} & \textbf{66.667}\\
    
    \midrule
    \rowcolor{gray!15}\multicolumn{9}{l}{Medium ($[6,15]$ objects)}\\
    \midrule
    Baseline & \underline{0.401} {\textbf{\scriptsize\textcolor{gray}{($\pm$0.02)}}} & \underline{6.933} {\textbf{\scriptsize\textcolor{gray}{($\pm$0.37)}}} & \underline{6.223} {\textbf{\scriptsize\textcolor{gray}{($\pm$0.59)}}} & \underline{6.199} {\textbf{\scriptsize\textcolor{gray}{($\pm$0.41)}}} & \underline{4.634} {\textbf{\scriptsize\textcolor{gray}{($\pm$0.56)}}} & \underline{5.997} {\textbf{\scriptsize\textcolor{gray}{($\pm$0.22)}}} & \underline{10.957} {\textbf{\scriptsize\textcolor{gray}{($\pm$2.11)}}} & \underline{17.143}\\
    Ours & \textbf{0.561} {\textbf{\scriptsize\textcolor{gray}{($\pm$0.02)}}} & \textbf{8.779} {\textbf{\scriptsize\textcolor{gray}{($\pm$0.18)}}} & \textbf{8.996} {\textbf{\scriptsize\textcolor{gray}{($\pm$0.23)}}} & \textbf{7.932} {\textbf{\scriptsize\textcolor{gray}{($\pm$0.24)}}} & \textbf{8.235} {\textbf{\scriptsize\textcolor{gray}{($\pm$0.29)}}} & \textbf{8.485} {\textbf{\scriptsize\textcolor{gray}{($\pm$0.13)}}} & \textbf{22.414} {\textbf{\scriptsize\textcolor{gray}{($\pm$19.15)}}} & \textbf{82.857}\\
    
    \midrule
    \rowcolor{gray!15}\multicolumn{9}{l}{Hard ($[16,20]$ objects)}\\
    \midrule
    Baseline & \underline{0.326} {\textbf{\scriptsize\textcolor{gray}{($\pm$0.02)}}} & \underline{6.808} {\textbf{\scriptsize\textcolor{gray}{($\pm$0.30)}}} & \underline{6.162} {\textbf{\scriptsize\textcolor{gray}{($\pm$0.59)}}} & \underline{6.099} {\textbf{\scriptsize\textcolor{gray}{($\pm$0.42)}}} & \underline{4.883} {\textbf{\scriptsize\textcolor{gray}{($\pm$0.58)}}} & \underline{5.988} {\textbf{\scriptsize\textcolor{gray}{($\pm$0.30)}}} & \underline{34.067} {\textbf{\scriptsize\textcolor{gray}{($\pm$14.89)}}} & \underline{6.667}\\
    Ours & \textbf{0.553} {\textbf{\scriptsize\textcolor{gray}{($\pm$0.03)}}} & \textbf{9.067} {\textbf{\scriptsize\textcolor{gray}{($\pm$0.13)}}} & \textbf{9.271} {\textbf{\scriptsize\textcolor{gray}{($\pm$0.17)}}} & \textbf{8.142} {\textbf{\scriptsize\textcolor{gray}{($\pm$0.27)}}} & \textbf{8.659} {\textbf{\scriptsize\textcolor{gray}{($\pm$0.25)}}} & \textbf{8.785} {\textbf{\scriptsize\textcolor{gray}{($\pm$0.07)}}} & \textbf{61.733} {\textbf{\scriptsize\textcolor{gray}{($\pm$28.80)}}} & \textbf{93.333}\\
    \bottomrule
    \end{tabular}%
    }
    \end{table*}
    
    \begin{table*}[tb]
    \centering
    \caption{\textbf{Per-Scene Quantitative Analysis.} We report the performance breakdown across 10 distinct scene themes. Our method demonstrates robust generalization, outperforming the baseline in nearly all metrics across diverse environments.}
    \label{tab:scene_gen:results_per_scene}
    \resizebox{\textwidth}{!}{%
    \begin{tabular}{llrrrrrrrr}
    \toprule
    \rowcolor{nvidiagreen!15}Theme & Method & VQA ($\uparrow$) & Real. ($\uparrow$) & Func. ($\uparrow$) & Lay. ($\uparrow$) & Compl. ($\uparrow$) & Qual. ($\uparrow$) & \# Obj ($\uparrow$) & GPT Pref. ($\uparrow$)\\
    \midrule
    
    \multirow{2}{*}{Bathroom Counter} 
    & Baseline & \underline{0.405} {\textbf{\scriptsize\textcolor{gray}{($\pm$0.02)}}} & \underline{6.901} {\textbf{\scriptsize\textcolor{gray}{($\pm$0.37)}}} & \underline{6.196} {\textbf{\scriptsize\textcolor{gray}{($\pm$0.73)}}} & \underline{6.260} {\textbf{\scriptsize\textcolor{gray}{($\pm$0.26)}}} & \underline{4.805} {\textbf{\scriptsize\textcolor{gray}{($\pm$0.56)}}} & \underline{6.041} {\textbf{\scriptsize\textcolor{gray}{($\pm$0.19)}}} & \underline{15.00} {\textbf{\scriptsize\textcolor{gray}{($\pm$0.00)}}} & \underline{10.00}\\
    & Ours & \textbf{0.564} {\textbf{\scriptsize\textcolor{gray}{($\pm$0.02)}}} & \textbf{8.913} {\textbf{\scriptsize\textcolor{gray}{($\pm$0.16)}}} & \textbf{9.159} {\textbf{\scriptsize\textcolor{gray}{($\pm$0.22)}}} & \textbf{7.967} {\textbf{\scriptsize\textcolor{gray}{($\pm$0.28)}}} & \textbf{8.276} {\textbf{\scriptsize\textcolor{gray}{($\pm$0.33)}}} & \textbf{8.579} {\textbf{\scriptsize\textcolor{gray}{($\pm$0.15)}}} & \textbf{28.50} {\textbf{\scriptsize\textcolor{gray}{($\pm$18.47)}}} & \textbf{90.00}\\
    \midrule
    
    \multirow{2}{*}{Classroom Supplies} 
    & Baseline & \underline{0.401} {\textbf{\scriptsize\textcolor{gray}{($\pm$0.02)}}} & \underline{6.881} {\textbf{\scriptsize\textcolor{gray}{($\pm$0.31)}}} & \underline{6.458} {\textbf{\scriptsize\textcolor{gray}{($\pm$0.50)}}} & \underline{5.931} {\textbf{\scriptsize\textcolor{gray}{($\pm$0.41)}}} & \underline{5.018} {\textbf{\scriptsize\textcolor{gray}{($\pm$0.55)}}} & \underline{6.072} {\textbf{\scriptsize\textcolor{gray}{($\pm$0.17)}}} & \underline{10.40} {\textbf{\scriptsize\textcolor{gray}{($\pm$1.26)}}} & \textbf{50.00}\\
    & Ours & \textbf{0.562} {\textbf{\scriptsize\textcolor{gray}{($\pm$0.02)}}} & \textbf{8.790} {\textbf{\scriptsize\textcolor{gray}{($\pm$0.18)}}} & \textbf{8.990} {\textbf{\scriptsize\textcolor{gray}{($\pm$0.18)}}} & \textbf{7.842} {\textbf{\scriptsize\textcolor{gray}{($\pm$0.28)}}} & \textbf{8.406} {\textbf{\scriptsize\textcolor{gray}{($\pm$0.26)}}} & \textbf{8.507} {\textbf{\scriptsize\textcolor{gray}{($\pm$0.12)}}} & \textbf{32.90} {\textbf{\scriptsize\textcolor{gray}{($\pm$23.70)}}} & \textbf{50.00}\\
    \midrule
    
    \multirow{2}{*}{Craft Station} 
    & Baseline & \underline{0.399} {\textbf{\scriptsize\textcolor{gray}{($\pm$0.02)}}} & \underline{7.062} {\textbf{\scriptsize\textcolor{gray}{($\pm$0.49)}}} & \underline{6.385} {\textbf{\scriptsize\textcolor{gray}{($\pm$0.56)}}} & \underline{6.171} {\textbf{\scriptsize\textcolor{gray}{($\pm$0.39)}}} & \underline{4.590} {\textbf{\scriptsize\textcolor{gray}{($\pm$0.60)}}} & \underline{6.052} {\textbf{\scriptsize\textcolor{gray}{($\pm$0.24)}}} & \underline{9.70} {\textbf{\scriptsize\textcolor{gray}{($\pm$0.48)}}} & \underline{10.00}\\
    & Ours & \textbf{0.560} {\textbf{\scriptsize\textcolor{gray}{($\pm$0.03)}}} & \textbf{8.761} {\textbf{\scriptsize\textcolor{gray}{($\pm$0.19)}}} & \textbf{9.045} {\textbf{\scriptsize\textcolor{gray}{($\pm$0.28)}}} & \textbf{7.938} {\textbf{\scriptsize\textcolor{gray}{($\pm$0.22)}}} & \textbf{8.179} {\textbf{\scriptsize\textcolor{gray}{($\pm$0.33)}}} & \textbf{8.481} {\textbf{\scriptsize\textcolor{gray}{($\pm$0.06)}}} & \textbf{17.70} {\textbf{\scriptsize\textcolor{gray}{($\pm$13.61)}}} & \textbf{90.00}\\
    \midrule
    
    \multirow{2}{*}{Garage Workstation} 
    & Baseline & \underline{0.410} {\textbf{\scriptsize\textcolor{gray}{($\pm$0.01)}}} & \underline{7.032} {\textbf{\scriptsize\textcolor{gray}{($\pm$0.34)}}} & \underline{6.308} {\textbf{\scriptsize\textcolor{gray}{($\pm$0.63)}}} & \underline{6.246} {\textbf{\scriptsize\textcolor{gray}{($\pm$0.47)}}} & \underline{4.432} {\textbf{\scriptsize\textcolor{gray}{($\pm$0.62)}}} & \underline{6.005} {\textbf{\scriptsize\textcolor{gray}{($\pm$0.26)}}} & \underline{9.00} {\textbf{\scriptsize\textcolor{gray}{($\pm$0.47)}}} & \underline{0.00}\\
    & Ours & \textbf{0.566} {\textbf{\scriptsize\textcolor{gray}{($\pm$0.02)}}} & \textbf{8.796} {\textbf{\scriptsize\textcolor{gray}{($\pm$0.21)}}} & \textbf{9.004} {\textbf{\scriptsize\textcolor{gray}{($\pm$0.20)}}} & \textbf{7.881} {\textbf{\scriptsize\textcolor{gray}{($\pm$0.28)}}} & \textbf{8.207} {\textbf{\scriptsize\textcolor{gray}{($\pm$0.29)}}} & \textbf{8.472} {\textbf{\scriptsize\textcolor{gray}{($\pm$0.13)}}} & \textbf{13.00} {\textbf{\scriptsize\textcolor{gray}{($\pm$11.68)}}} & \textbf{100.00}\\
    \midrule
    
    \multirow{2}{*}{Garden Tools} 
    & Baseline & \underline{0.400} {\textbf{\scriptsize\textcolor{gray}{($\pm$0.02)}}} & \underline{7.018} {\textbf{\scriptsize\textcolor{gray}{($\pm$0.45)}}} & \underline{6.155} {\textbf{\scriptsize\textcolor{gray}{($\pm$0.53)}}} & \underline{6.315} {\textbf{\scriptsize\textcolor{gray}{($\pm$0.47)}}} & \underline{4.489} {\textbf{\scriptsize\textcolor{gray}{($\pm$0.53)}}} & \underline{5.994} {\textbf{\scriptsize\textcolor{gray}{($\pm$0.21)}}} & \underline{10.30} {\textbf{\scriptsize\textcolor{gray}{($\pm$0.95)}}} & \underline{30.00}\\
    & Ours & \textbf{0.561} {\textbf{\scriptsize\textcolor{gray}{($\pm$0.02)}}} & \textbf{8.778} {\textbf{\scriptsize\textcolor{gray}{($\pm$0.16)}}} & \textbf{8.949} {\textbf{\scriptsize\textcolor{gray}{($\pm$0.15)}}} & \textbf{7.950} {\textbf{\scriptsize\textcolor{gray}{($\pm$0.20)}}} & \textbf{8.175} {\textbf{\scriptsize\textcolor{gray}{($\pm$0.27)}}} & \textbf{8.463} {\textbf{\scriptsize\textcolor{gray}{($\pm$0.12)}}} & \textbf{13.80} {\textbf{\scriptsize\textcolor{gray}{($\pm$11.36)}}} & \textbf{70.00}\\
    \midrule
    
    \multirow{2}{*}{Kitchen Cabinet} 
    & Baseline & \underline{0.327} {\textbf{\scriptsize\textcolor{gray}{($\pm$0.02)}}} & \underline{6.862} {\textbf{\scriptsize\textcolor{gray}{($\pm$0.31)}}} & \underline{6.281} {\textbf{\scriptsize\textcolor{gray}{($\pm$0.61)}}} & \underline{6.008} {\textbf{\scriptsize\textcolor{gray}{($\pm$0.39)}}} & \underline{5.069} {\textbf{\scriptsize\textcolor{gray}{($\pm$0.48)}}} & \underline{6.055} {\textbf{\scriptsize\textcolor{gray}{($\pm$0.26)}}} & \underline{37.38} {\textbf{\scriptsize\textcolor{gray}{($\pm$19.66)}}} & \underline{12.50}\\
    & Ours & \textbf{0.554} {\textbf{\scriptsize\textcolor{gray}{($\pm$0.03)}}} & \textbf{9.052} {\textbf{\scriptsize\textcolor{gray}{($\pm$0.13)}}} & \textbf{9.313} {\textbf{\scriptsize\textcolor{gray}{($\pm$0.17)}}} & \textbf{8.070} {\textbf{\scriptsize\textcolor{gray}{($\pm$0.25)}}} & \textbf{8.710} {\textbf{\scriptsize\textcolor{gray}{($\pm$0.25)}}} & \textbf{8.786} {\textbf{\scriptsize\textcolor{gray}{($\pm$0.06)}}} & \textbf{48.88} {\textbf{\scriptsize\textcolor{gray}{($\pm$25.63)}}} & \textbf{87.50}\\
    \midrule
    
    \multirow{2}{*}{Laundry Sorting} 
    & Baseline & \underline{0.396} {\textbf{\scriptsize\textcolor{gray}{($\pm$0.01)}}} & \underline{6.795} {\textbf{\scriptsize\textcolor{gray}{($\pm$0.24)}}} & \underline{6.327} {\textbf{\scriptsize\textcolor{gray}{($\pm$0.63)}}} & \underline{6.269} {\textbf{\scriptsize\textcolor{gray}{($\pm$0.34)}}} & \underline{4.536} {\textbf{\scriptsize\textcolor{gray}{($\pm$0.47)}}} & \underline{5.982} {\textbf{\scriptsize\textcolor{gray}{($\pm$0.21)}}} & \underline{12.30} {\textbf{\scriptsize\textcolor{gray}{($\pm$1.70)}}} & \underline{0.00}\\
    & Ours & \textbf{0.558} {\textbf{\scriptsize\textcolor{gray}{($\pm$0.03)}}} & \textbf{8.742} {\textbf{\scriptsize\textcolor{gray}{($\pm$0.17)}}} & \textbf{8.975} {\textbf{\scriptsize\textcolor{gray}{($\pm$0.26)}}} & \textbf{8.021} {\textbf{\scriptsize\textcolor{gray}{($\pm$0.16)}}} & \textbf{8.202} {\textbf{\scriptsize\textcolor{gray}{($\pm$0.28)}}} & \textbf{8.485} {\textbf{\scriptsize\textcolor{gray}{($\pm$0.12)}}} & \textbf{35.30} {\textbf{\scriptsize\textcolor{gray}{($\pm$27.34)}}} & \textbf{100.00}\\
    \midrule
    
    \multirow{2}{*}{Office Desk} 
    & Baseline & \underline{0.457} {\textbf{\scriptsize\textcolor{gray}{($\pm$0.02)}}} & \underline{6.747} {\textbf{\scriptsize\textcolor{gray}{($\pm$0.36)}}} & \underline{6.183} {\textbf{\scriptsize\textcolor{gray}{($\pm$0.25)}}} & \underline{6.101} {\textbf{\scriptsize\textcolor{gray}{($\pm$0.52)}}} & \underline{4.697} {\textbf{\scriptsize\textcolor{gray}{($\pm$0.59)}}} & \underline{5.932} {\textbf{\scriptsize\textcolor{gray}{($\pm$0.25)}}} & \underline{7.00} {\textbf{\scriptsize\textcolor{gray}{($\pm$0.00)}}} & \textbf{57.14}\\
    & Ours & \textbf{0.499} {\textbf{\scriptsize\textcolor{gray}{($\pm$0.05)}}} & \textbf{8.296} {\textbf{\scriptsize\textcolor{gray}{($\pm$0.22)}}} & \textbf{8.535} {\textbf{\scriptsize\textcolor{gray}{($\pm$0.16)}}} & \textbf{7.524} {\textbf{\scriptsize\textcolor{gray}{($\pm$0.16)}}} & \textbf{7.609} {\textbf{\scriptsize\textcolor{gray}{($\pm$0.32)}}} & \textbf{7.991} {\textbf{\scriptsize\textcolor{gray}{($\pm$0.08)}}} & \textbf{17.57} {\textbf{\scriptsize\textcolor{gray}{($\pm$16.03)}}} & \underline{42.86}\\
    \midrule
    
    \multirow{2}{*}{Storage Room} 
    & Baseline & \underline{0.324} {\textbf{\scriptsize\textcolor{gray}{($\pm$0.02)}}} & \underline{6.747} {\textbf{\scriptsize\textcolor{gray}{($\pm$0.29)}}} & \underline{6.027} {\textbf{\scriptsize\textcolor{gray}{($\pm$0.58)}}} & \underline{6.203} {\textbf{\scriptsize\textcolor{gray}{($\pm$0.45)}}} & \underline{4.670} {\textbf{\scriptsize\textcolor{gray}{($\pm$0.64)}}} & \underline{5.912} {\textbf{\scriptsize\textcolor{gray}{($\pm$0.35)}}} & \underline{30.29} {\textbf{\scriptsize\textcolor{gray}{($\pm$5.91)}}} & \underline{0.00}\\
    & Ours & \textbf{0.552} {\textbf{\scriptsize\textcolor{gray}{($\pm$0.02)}}} & \textbf{9.085} {\textbf{\scriptsize\textcolor{gray}{($\pm$0.14)}}} & \textbf{9.224} {\textbf{\scriptsize\textcolor{gray}{($\pm$0.17)}}} & \textbf{8.224} {\textbf{\scriptsize\textcolor{gray}{($\pm$0.28)}}} & \textbf{8.601} {\textbf{\scriptsize\textcolor{gray}{($\pm$0.24)}}} & \textbf{8.783} {\textbf{\scriptsize\textcolor{gray}{($\pm$0.08)}}} & \textbf{76.43} {\textbf{\scriptsize\textcolor{gray}{($\pm$26.40)}}} & \textbf{100.00}\\
    \midrule
    
    \multirow{2}{*}{Tea Time} 
    & Baseline & \underline{0.458} {\textbf{\scriptsize\textcolor{gray}{($\pm$0.02)}}} & \underline{6.784} {\textbf{\scriptsize\textcolor{gray}{($\pm$0.39)}}} & \underline{6.345} {\textbf{\scriptsize\textcolor{gray}{($\pm$0.77)}}} & \underline{6.061} {\textbf{\scriptsize\textcolor{gray}{($\pm$0.48)}}} & \underline{4.773} {\textbf{\scriptsize\textcolor{gray}{($\pm$0.67)}}} & \underline{5.991} {\textbf{\scriptsize\textcolor{gray}{($\pm$0.29)}}} & \underline{6.00} {\textbf{\scriptsize\textcolor{gray}{($\pm$0.00)}}} & \underline{12.50}\\
    & Ours & \textbf{0.547} {\textbf{\scriptsize\textcolor{gray}{($\pm$0.03)}}} & \textbf{8.361} {\textbf{\scriptsize\textcolor{gray}{($\pm$0.26)}}} & \textbf{8.325} {\textbf{\scriptsize\textcolor{gray}{($\pm$0.31)}}} & \textbf{7.735} {\textbf{\scriptsize\textcolor{gray}{($\pm$0.21)}}} & \textbf{7.642} {\textbf{\scriptsize\textcolor{gray}{($\pm$0.18)}}} & \textbf{8.016} {\textbf{\scriptsize\textcolor{gray}{($\pm$0.10)}}} & \textbf{8.63} {\textbf{\scriptsize\textcolor{gray}{($\pm$1.85)}}} & \textbf{87.50}\\
    \midrule
    
    \multirow{2}{*}{Workshop Bench} 
    & Baseline & \underline{0.394} {\textbf{\scriptsize\textcolor{gray}{($\pm$0.02)}}} & \underline{6.838} {\textbf{\scriptsize\textcolor{gray}{($\pm$0.35)}}} & \underline{5.733} {\textbf{\scriptsize\textcolor{gray}{($\pm$0.38)}}} & \underline{6.199} {\textbf{\scriptsize\textcolor{gray}{($\pm$0.47)}}} & \underline{4.564} {\textbf{\scriptsize\textcolor{gray}{($\pm$0.52)}}} & \underline{5.833} {\textbf{\scriptsize\textcolor{gray}{($\pm$0.22)}}} & \underline{10.00} {\textbf{\scriptsize\textcolor{gray}{($\pm$0.47)}}} & \underline{20.00}\\
    & Ours & \textbf{0.556} {\textbf{\scriptsize\textcolor{gray}{($\pm$0.03)}}} & \textbf{8.675} {\textbf{\scriptsize\textcolor{gray}{($\pm$0.10)}}} & \textbf{8.846} {\textbf{\scriptsize\textcolor{gray}{($\pm$0.22)}}} & \textbf{7.928} {\textbf{\scriptsize\textcolor{gray}{($\pm$0.26)}}} & \textbf{8.200} {\textbf{\scriptsize\textcolor{gray}{($\pm$0.30)}}} & \textbf{8.412} {\textbf{\scriptsize\textcolor{gray}{($\pm$0.16)}}} & \textbf{15.70} {\textbf{\scriptsize\textcolor{gray}{($\pm$10.36)}}} & \textbf{80.00}\\
    
    \bottomrule
    \end{tabular}%
    }
    \end{table*}

\section{Details on Task Generation Evaluation}
\label{sec:task-generation-evaluation}

We evaluated the quality of our task generation method using an LLM-as-judge framework.
Tasks were generated by prompting an LLM (o1~\cite{openai2024openaio1card}) with scene descriptions and our Category templates\footnote{For 2 simple scenes, we generated 1 task for each of the 7 categories and for the remaining 57 scenes, we generated 2 tasks. This produced $57*7*2 + 2*7*1 = 812$ tasks}.
For each generated task, we extracted the natural language instruction and the corresponding termination conditions (success criteria implemented as predicate functions) through static analysis of the generated Python code.
We then prompted an LLM (also o1~\cite{openai2024openaio1card}) to assess alignment between the instruction and the programmatic success conditions across six dimensions: relation match (whether the spatial/logical relationship is preserved), target match (correctness of goal state), object match (whether referenced objects are correct), quantifier match (handling of ``all,'' ``any,'' or specific counts), instruction clarity (unambiguous and well-formed language), and physical feasibility (whether the task is achievable given typical robot capabilities).
Each dimension was scored on a 0–1 scale, and we computed an aggregate alignment score as the weighted mean.
The model additionally provided a categorical verdict—aligned, partially aligned, or misaligned—based on whether the termination conditions would correctly evaluate task success as described in the instruction.

\begin{table*}[t]
\centering
\caption{LLM-judged quality metrics for $812$ automatically generated manipulation tasks across $59$ scenes and $7$ competency axes.}
\label{tab:task-eval}
\vspace{-0.5em}
\begin{tabular}{@{}lr cccccc cc@{}}
\toprule
\rowcolor{nvidiagreen!15}& & \multicolumn{6}{c}{\textbf{LLM Judge}} & \multicolumn{2}{c}{\textbf{Coverage}} \\
\cmidrule(lr){3-8} \cmidrule(lr){9-10}
\rowcolor{nvidiagreen!15}Category & $N$ & Alignment & Clarity & Feasibility & Match & Aligned\% & Partial\% & Object & Predicate \\
\midrule
color & 116 & 0.81 & 0.94 & 0.80 & 0.90 & 57 & 40 & 0.88 & 0.29 \\
conjunction & 116 & 0.97 & 0.98 & 1.00 & 0.98 & 91 & 9 & 0.88 & 0.29 \\
counting & 116 & 0.87 & 0.97 & 0.90 & 0.92 & 60 & 38 & 0.88 & 0.29 \\
recognition & 116 & 0.96 & 0.97 & 0.96 & 0.97 & 85 & 15 & 0.88 & 0.29 \\
semantics & 116 & 0.89 & 0.95 & 0.94 & 0.94 & 72 & 27 & 0.88 & 0.29 \\
sorting & 116 & 0.94 & 0.95 & 0.97 & 0.96 & 86 & 14 & 0.88 & 0.29 \\
spatial & 116 & 0.92 & 0.98 & 0.89 & 0.95 & 80 & 17 & 0.88 & 0.29 \\
\midrule
\rowcolor{gray!15}\textbf{Overall} & \textbf{812} & \textbf{0.91} & \textbf{0.96} & \textbf{0.92} & \textbf{0.95} & \textbf{76} & \textbf{23} & \textbf{0.88} & \textbf{0.29} \\
\bottomrule
\end{tabular}
\vspace{-1em}
\end{table*}

Table~\ref{tab:task-eval} shows that our method can successfully generate a variety of types of tasks appropriate to the assets in the scene.
The \textbf{Alignment} score represents the overall instruction-code alignment, aggregating the six sub-dimensions.
\textbf{Clarity} measures whether instructions are unambiguous and grammatically well-formed.
\textbf{Feasibility} assesses physical realizability of the task.
\textbf{Match} combines the four semantic dimensions (relation, target, object, quantifier) into a single score reflecting how accurately the code captures the instruction's intent.
\textbf{Verdict} reports the percentage of tasks judged as fully aligned versus partially aligned (misaligned tasks, comprising approximately 1\% of the dataset, are omitted for brevity).
We additionally compute scene coverage metrics: object coverage measures the fraction of manipulable objects in each scene that appear in at least one generated task, while predicate coverage measures the fraction of available termination predicates used across tasks for that scene.
All evaluations use temperature 0 for reproducibility, with automatic retry logic to handle rate limits.

The evaluation reveals strong overall task generation quality, with $0.91$ mean alignment and $76\%$ of tasks receiving full alignment verdicts. Performance varies by category: conjunction and recognition tasks achieve the highest alignment ($0.97$ and $0.96$), likely because their success conditions map straightforwardly to compositional predicates, while color-based tasks show lower alignment ($0.81$), reflecting the challenge of grounding color references to specific object instances. High clarity ($0.96$) and semantic match ($0.95$) scores indicate that the generated instructions are well-formed and the termination conditions capture the intended semantics, though feasibility scores are slightly lower for spatial tasks ($0.89$) where precise placement requirements may exceed typical manipulation tolerances. The $88\%$ object coverage demonstrates good utilization of scene assets, while the lower predicate coverage ($29\%$) suggests the generator favors a subset of reliable predicates rather than exploring the full space of available success conditions---a conservative strategy that likely contributes to the high alignment scores.
Overall, these results demonstrate our method can successfully generate a variety of types of tasks appropriate to the assets in the scene.
While LLM-generated scenes and tasks are procedurally checked, we note that scenes and tasks require user review in order to ensure that they are aligned with the evaluation objective.

\begin{algorithm}[tb]
    \caption{Physical Placement Solver}
    \label{alg:physical}
    \textbf{Input:} Objects $B$, Predicates $P$, Solved Base Poses \\
    \textbf{Output:} 3D coordinates $(x, y, z)$ for all objects
    \begin{algorithmic}[1]
    \STATE \COMMENT{\textcolor{blue}{Solve Stacking:}}
    \FORALL{$p \in P \text{ where } p.\text{type} == \text{place-on}$}
        \STATE $s \gets p.\text{support}$
        \STATE $B_{\text{peers}} \gets \{b' \mid b' \text{ is already on } s\}$
        \STATE $(x, y) \gets \text{FindSpot}(s, p.\text{object}, B_{\text{peers}})$
        \STATE $p.\text{object}.z \gets s.z + s.\text{height} + p.\text{object}.\text{height}/2$
        \STATE $p.\text{object}.(x,y) \gets (x,y)$
    \ENDFOR
    \STATE \COMMENT{\textcolor{blue}{Solve Containment:}}
    \FORALL{$p \in P \text{ where } p.\text{type} == \text{place-in}$}
        \STATE $c \gets p.\text{container}$
        \IF{$\text{TotalArea}(p.\text{objects}) > 0.8 \times \text{Area}(c)$}
            \STATE $p.\text{objects} \gets \text{SortAndFilter}(p.\text{objects}, c.\text{capacity})$
        \ENDIF
        \STATE $(R, C) \gets \text{ComputeGridDimensions}(c.\text{dims}, |p.\text{objects}|)$
        \FOR{$i = 0$ \textbf{to} $|p.\text{objects}|-1$}
            \STATE $(r, c) \gets (i // C, i \% C)$
            \STATE $(x_{\text{loc}}, y_{\text{loc}}) \gets \text{GridCellCenter}(r, c, c.\text{dims})$
            \STATE $\text{Jitter}(x_{\text{loc}}, y_{\text{loc}})$
            \STATE $p.\text{objects}[i].(x,y) \gets c.(x,y) + (x_{\text{loc}}, y_{\text{loc}})$
            \STATE $p.\text{objects}[i].z \gets c.z + c.\text{height}/2 + \text{buffer}$
        \ENDFOR
    \ENDFOR
    \RETURN \textbf{Success}
    \end{algorithmic}
\end{algorithm}

\end{document}